\documentclass[accepted]{uai2022} 

\usepackage[american]{babel}

\usepackage{natbib} 
\usepackage{mathtools} 
\usepackage{booktabs} 
\usepackage{tikz}
\usepackage{multirow}

\usepackage{graphicx}
\usepackage{subcaption}
\usepackage{algorithm}
\usepackage{algorithmic}

\usepackage{amsmath}
\usepackage{amssymb}
\usepackage{mathtools}
\usepackage{amsthm}

\usepackage{bm}
\usepackage[inline=true, margin=false]{fixme}

\theoremstyle{plain}
\newtheorem{theorem}{Theorem}[section]

\theoremstyle{definition}

\theoremstyle{remark}

\title{A Mutually Exciting Latent Space Hawkes Process Model\\
for Continuous-time Networks}

\author[1]{\href{mailto:<zhipeng.huang@rockets.utoledo.edu>?Subject=Your UAI 2022 paper}{Zhipeng Huang}{}}
\author[1]{Hadeel Soliman}
\author[2]{Subhadeep Paul}
\author[1]{Kevin S. Xu}

\affil[1]{%
    Department of Electrical Engineering and Computer Science\\
    University of Toledo\\
    Toledo, OH, USA
}
\affil[2]{%
    Department of Statistics\\
    The Ohio State University\\
    Columbus, OH, USA
}

\begin{document}
\maketitle

\begin{abstract}
Networks and temporal point processes serve as fundamental building blocks for modeling complex dynamic relational data in various domains. 
We propose the \emph{latent space Hawkes (LSH)} model, a novel generative model for continuous-time networks of relational events, using a latent space representation for nodes. 
We model relational events between nodes using mutually exciting Hawkes processes with baseline intensities dependent upon the distances between the nodes in the latent space and sender and receiver specific effects. 
We demonstrate that our proposed LSH model can replicate many features observed in real temporal networks including reciprocity and transitivity,  while also achieving superior prediction accuracy and providing more interpretable fits than existing models.
\end{abstract}

\section{Introduction}
Dynamic networks are used to represent time-varying relationships (edges) between a set of nodes. 
They are useful in a variety of application settings, including 
messages between users on online social networks and transactions between users on online marketplaces. 
In such settings, the network typically evolves over time through a set of \emph{timestamped relational events}. Each event is a triplet $(u,v,t)$ denoting that node $u$ initiated an interaction with node $v$ (e.g.~$u$ sent a message to $v$) at timestamp $t$. 
We refer to this type of dynamic network as a \emph{continuous-time network} because it is continuously evolving through these relational events.

A topic of much recent interest is identifying latent representations for nodes in networks. 
These latent representations are often referred to as node embeddings, and node embedding-based approaches for common network analysis tasks including link prediction have gained significant attention in recent years \citep{grover2016node2vec, goyal2018graph, cui2018survey}. 
Prior to this surge of interest, latent space models have been used in statistics and mathematical sociology for exploratory analysis of networks \citep{hoff2002latent, hoff2005bilinear, hoff2007modeling, handcock2007model, krivitsky2009representing}. 
Latent representations have also been developed for dynamic networks evolving over discrete time steps  \citep{sewell2015latent} or in continuous time \citep{nguyen2018continuous}. 

Latent space representations can be combined with temporal point processes (TPPs) to form a probabilistic generative model for continuous-time networks, which we consider in this paper. 
Augmenting the latent representation with a TPP enables one to generate timestamps for the edges between nodes. 
\citet{yang2017decoupling} proposed the dual latent space (DLS) generative model that combines two types of latent spaces with bivariate Hawkes processes.
They found that using two types of latent spaces, one to capture homophily and one to capture reciprocity, provides a richer model that also leads to improved link prediction accuracy. 
However, much of the interpretability of the latent space, which was the original motivation of the latent space model of \citet{hoff2002latent}, is lost by using multiple high-dimensional latent spaces. 
Furthermore, the DLS model has issues with stability of the generative process due to the multiple latent spaces. 
It also uses only reciprocal excitation and not self excitation. 
Self excitation is important in application settings such as modeling text messages, where person $u$ may send multiple messages to $v$ in rapid succession before $v$ responds.

In this paper, we consider using a single latent space representation to provide a more interpretable model. 
The single latent space limits the flexibility of the model compared to the DLS, so we increase flexibility by adding self excitation and sender and receiver effects.
We demonstrate that our proposed latent space Hawkes (LSH) model is competitive with other models in predictive and generative tasks on 4 real network datasets while providing more interpretable and stable model fits.
Furthermore, we apply our LSH model to perform exploratory analysis on a dataset of militarized disputes to reveal network structure between countries.

\section{Background}

\subsection{Hawkes Processes}
\label{sec:background_hawkes}

The Hawkes process model was introduced for temporal point processes by \citet{hawkes1971spectra}. The defining characteristic of a Hawkes process is that it is self exciting, meaning that each event increases the rate of future events for some period of time. Mutually exciting Hawkes processes allow events from different processes to excite each other in addition to self excitation \citep{laub2021elements}. 
An $m$-dimensional mutually exciting Hawkes process is characterized by a conditional intensity function for each dimension $i$:
\begin{equation}
\label{eq:1}
\lambda_i^*(t) = \lambda_i (t|\mathcal{H}_t) 
= \mu_i + \sum_{j=1}^m \sum_{k:t_k < t} \phi_{ij}(t-t_k),
\end{equation}
where $\mathcal{H}_t$ denotes the history of the process up to time $t$,  $\mu_i$ denotes the baseline rate of events in dimension $i$, and $\phi_{ij}(\cdot)$ is a kernel function that describes how an event in dimension $j$ influences dimension $i$. 

The most commonly used kernel function is the exponential kernel $\phi(t-t_k)=\alpha \beta e^{-\beta(t-t_k)}$ for $\alpha > 0$ and $\beta > 0$. With each event arrival, the conditional intensity jumps by $\alpha$. The influence of the arrival then exponentially decays at rate $\beta$ over time. In practice, both $\alpha$ and $\beta$ are unknown parameters that need to be estimated from data, which is usually done using maximum likelihood estimation \citep{laub2021elements}. However, estimators for the decay parameter $\beta$ are poorly behaved \citep{santos2021surfacing}, and 
it is more computationally efficient to choose a fixed $\beta$ rather than estimating it \citep{lemonnier2014nonparametric}. 

An approach that is more general than fixing the value of $\beta$ is the sum of exponential kernels method \citep{lemonnier2014nonparametric}, which defines $\phi(t-t_k)= \sum^B_b \alpha \beta_b e^{-\beta_b(t-t_k)}$, where $B$ denotes the number of exponential kernels. This method generalizes better as it handles different time scales, which makes the modeling less sensitive to choice of $\beta$. 
We use the sum of exponential kernels decay in this paper.

\subsection{Latent Space Models}
\label{sec:LSMs}

The latent space model (LSM), first proposed by \citet{hoff2002latent} is a popular model-based approach for  social network analysis. 
Designed initially for a single static undirected network, the LSM allows the probability of an edge between two nodes to depend on their Euclidean distance in an unobserved or latent space using a logistic regression model. 
Let $A$ denote the adjacency matrix of a network, with $a_{uv} = 1$ for node pairs $(u,v)$ with an edge and $a_{uv} = 0$ otherwise.
By assuming conditional independence between node pairs, the log-likelihood can be written as
\begin{equation*}
\log P(A|\eta) = \sum_{u < v} \left[ \eta_{uv} a_{uv} - \log(1 + e^{\eta_{uv}}) \right],
\end{equation*}
where entry $\eta_{uv}$ in the matrix $\eta$ denotes the log odds of an edge being formed between nodes $(u,v)$. 
$\eta_{uv}$ is parameterized as follows:
$\eta_{uv} = \xi - \|z_u - z_v\|_2$,
where $z_u$ denotes the latent position of node $u$ in a $d$-dimensional latent space, and $\xi$ is an intercept term. 
Under this parameterization, two nodes with closer latent positions have higher probability of forming an edge.

The latent space model provides a visual and interpretable model-based spatial representation of social relationships. 
It has been extended by many researchers. \citet{handcock2007model} developed a latent position cluster model to capture transitivity, homophily, and community structure simultaneously. The latent space models were later extended to include node-specific random effects by  \citet{krivitsky2009representing}. 
Latent space models have also been extended for more complex network based data structures, including multiple networks \citep{gollini2016joint,salter2017latent}, discrete-time dynamic networks \citep{sewell2015latent,sewell2016latent, friel2016interlocking, gracious2021neural}, and multimodal networks \citep{wang2019joint}.
We use the latent space model as the building block for our proposed continuous-time LSH model.

\subsection{Related Work}

\paragraph{Dynamic Network Embeddings}

One line of related work is focused on node embeddings for dynamic networks. 
Compared to static network embedding methods, dynamic network embedding methods assign nodes low-dimensional representations that effectively preserve the temporal information. \citet{nguyen2018continuous} proposed continuous-time dynamic network embeddings (CTDNE), a general framework to learn a time-respecting embedding from continuous-time dynamic networks. Their framework acts as a basis for incorporating temporal dependencies into existing node embedding and deep graph models based on random walks. 
Other approaches for dynamic network embedding have also been proposed \citep{chen2018scalable, sankar2018dynamic, goyal2020dyngraph2vec}, many of which are discussed in a recent survey on dynamic network embedding \citep{xie2020survey}.

\paragraph{TPP-based Network Models}
TPP-based network models are generative models for continuous-time dynamic networks that incorporate both a generative process for the nodes $(u,v)$ that form an edge and the time $t$ at which an edge is formed. 
These timestamped edges or events can be viewed as triplets $(u,v,t)$. 
Many TPP-based network models utilize a discrete latent variable representation for the nodes \citep{Blundell2012,Dubois2013,miscouridou2018modelling,junuthula2019block,arastuie2020chip, soliman2022multivariate}, dividing them into different blocks or communities.

The most closely related work to this paper is the dual latent space (DLS) model \citep{yang2017decoupling}, which also utilizes a continuous latent variable representation inspired by the latent space model. 
The DLS model uses bivariate Hawkes processes to capture the homophily and reciprocity of dynamic networks. 
They observed that the latent dimensions of users which affect link formation may be different from the latent dimensions of users which affect reciprocity. 
We discuss shortcomings of the DLS model and its relation to our proposed model in Section \ref{sec:relation_DLS}.

Another TPP-based network model using a continuous latent space is proposed by \citet{rastelli2021continuous}. 
It assumes that the latent positions of nodes may change at a set of predefined change points rather than being fixed over time.

\paragraph{Other Continuous-time Network Models}
Earlier research on continuous-time network models was proposed by \citet{wasserman1980analyzing, wasserman1980stochastic}, who modeled the evolution of network data using continuous-time Markov chains. 
Later on, \citet{snijders2005models, snijders2017modeling} proposed a set of network models that offers more flexibility to represent a variety of network effects, such as transitivity, reciprocity, etc. \citet{fan2012learning} explored the inference for these models and proposed a sampling-based learning algorithm for continuous-time social network models.

\section{Proposed Model}

In our model, we employ a latent space to learn hidden node attributes underlying the network and mutually exciting Hawkes processes to capture the temporal dynamics of communication. We model the communications between each pair of nodes as realizations from a bivariate Hawkes process whose conditional intensity function $\lambda_{uv}(t|\mathcal{H}_t)$ includes three components: a baseline rate, a self-exciting term, and a reciprocal term. 

Let $z_u$ and $z_v$ denote the latent positions for nodes $u$ and $v$, respectively. We model baseline rate $\mu_{uv}$ as a function of Euclidean distances between $z_u$ and $z_v$. \citet{gollini2016joint} showed that  squared Euclidean distances are computationally more efficient than  Euclidean distances yet resulted in similar latent positions. Thus, we use squared Euclidean distances $||z_u - z_v||_2^2$ in the model for $\mu_{uv}$, similar to DLS \citep{yang2017decoupling}. 
We further add sender and receiver node effect terms $\delta_u, \gamma_v$ to the model as in \cite{hoff2005bilinear,krivitsky2009representing,wang2019joint} to capture the degree heterogeneity, namely the tendency of some nodes to send and receive events more than others, respectively. 

A Hawkes process with exponential kernel has been found to be a good model for conversation event sequences as well as other relational temporal event data \citep{masuda2013self}. 
We use a sum of $B$ exponential kernels in our Hawkes processes. 
We set $\beta = (\beta_1, \beta_2, \ldots, \beta_B)$ as a set of fixed known decays and $C = (C_1, C_2, \ldots,C_B)$ as a set of scaling parameters for the kernel with $\sum_i^B C_i= 1$. 
The conditional intensity function can be written as follows:
\begin{equation} \label{eq:lsh_lambda}
\begin{split}
\lambda_{uv}^*(t) &= \mu_{uv} +   \sum_{t_{uv} < t}\sum_b^B C_b \alpha_1 \beta_b e^{-\beta_b(t-t_{uv})} \\
&+   \sum_{t_{vu} < t}\sum_b^B C_b \alpha_2 \beta_b e^{-\beta_b(t-t_{vu})}, \quad \forall u \neq v
\end{split}
\end{equation}
where the baseline rate $\mu_{uv}$ is given by
\begin{equation}
\label{eq:baseline}
\mu_{uv} = e^{-\theta_1||z_u - z_v||^2_2 + \theta_2 + \delta_u + \gamma_v}.  
\end{equation}

\subsection{Model Parameters}
The LSH model has parameters ($Z$, $\alpha_1$, $\alpha_2$, $\theta_1$, $\theta_2$ $\delta$, $\gamma$). Each node has a $d$-dimensional latent position $z_u$, a sender propensity parameter $\delta_u$ and a receiver propensity parameter $\gamma_u$. $\alpha_1$ and $\alpha_2$ are the jump size parameters for self-excitation and reciprocal-excitation. $Z$ is a $n \times d$ matrix where each row is  a latent position vector $z_u$ for a node, and $d$ is the latent dimension. Each of $\delta$ and $\gamma$ is a vector of size $n$. 
$\theta_1$ and $\theta_2$ are slope and intercept terms, respectively, associated with the baseline rate and latent positions. A positive slope $\theta_1$ provides node pairs closer together in the latent space with a higher probability of forming edges, while a negative slope does the reverse.

\paragraph{Identifiability}
There are two sets of identifiability problems that need to be discussed. From the observed event times, the Hawkes process parameters $\mu_{uv}, \alpha_1, \alpha_2$ can be identified as shown by \citet{ozaki1979maximum}. With the baseline intensity parameter $\mu_{uv}$ correctly identified, we explore the identifiability of the parameters in the model for $\mu_{uv}$. The identifiability of parameters in the latent space model has been discussed by \cite{ma2020universal} for a single network and by \cite{zhang2020flexible} for multilayer networks. 

Denote $1_n$ to be the $n$ dimensional vector and $J_n=1_n 1_n^T$ to be the $n\times n$ matrix whose elements are all 1's. We first note that the magnitude of the parameter $\theta_1$ is not identifiable since it enters the equation for $\mu$ as a product with $\|z_u - z_v\|^2_2$. 
However, the sign of $\theta_1$ is identifiable since $\|z_u - z_v\|^2_2$ is always positive. In the following, we set $\theta_1=1$ and examine the conditions for identification of other parameters.
We have
\begin{align*}
\log (\mu_{uv}) &  = \theta_2 - \|z_u\|^2 - \|z_v\|^2 + z_u^Tz_v + \delta_u +\gamma_v \\
& = \theta_2 + z_u^Tz_v + \tilde{\delta}_u +\tilde{\gamma}_v,
\end{align*}
where $\tilde{\delta}_u = \delta_u - \|z_u\|^2$ and $\tilde{\gamma}_v = \gamma_v - \|z_v\|^2$.
Now  let $\tilde{\delta}$ and $\tilde{\gamma}$ denote the $n$-dimensional vectors whose elements are $\tilde{\delta}_u$ and $\tilde{\gamma}_v$, respectively. (All vectors are column vectors.) Writing in matrix form, the above expression is 
\begin{equation*}
\log (\mu) = \theta_2 J_n + ZZ^T + \tilde{\delta}1_n^T + 1_n \tilde{\gamma}^T. 
\end{equation*}

\begin{theorem}
\label{thm:identifiability}
Under the following assumptions:
\begin{enumerate}
    \item The latent positions are centered, i.e., $HZ=Z$, where $H=I-\frac{1}{n}11^T$, and
    \item The total nodal effects sum to 0, i.e., $1_n^T \tilde{\delta}=0$ and $\tilde{\gamma}^T1_n = 0$,
\end{enumerate}
if two sets of parameters $\theta_2, Z,\gamma,\delta$ and $\theta_2', Z',\gamma',\delta'$ lead to the same $\log (\mu)$, then
\begin{equation*}
\theta_2 =\theta_2', \,\, \delta = \delta', \,\, \gamma=\gamma'\,\, \text{ and }  Z =Z'O,
\end{equation*}
where $O$ is a $d \times d$ orthogonal matrix.
\label{identifiability}
\end{theorem}
The proof is provided in Appendix \ref{sec:supp_proof}. 
Thus, under the constraints that the true latent positions $Z$ are centered and total nodal sender and receiver effects sum to 0, the parameters $\theta_2$, $\delta$, $\gamma$ and the vector distances $ZZ^T$ are exactly identified, while $Z$ is identified up to an orthogonal matrix $O$.

\subsection{Relation to DLS Model}
\label{sec:relation_DLS}

The most similar model to ours is the dual latent space (DLS) model \citep{yang2017decoupling}. 
It uses the following form for the conditional intensity function\footnote{They include also a periodic kernel in addition to the exponential kernels, which we exclude for ease of comparison.}: 
\begin{equation} \label{eq:dls_lambda}
\begin{split}
&\lambda_{uv}^*(t) = e^{-||z_u - z_v||^2_2 + \theta_2} \\
&+   \sum_{t_{vu} < t}\sum_b^B \alpha_2 e^{-||x_u^{(b)} - x_v^{(b)}||^2_2} \beta_b e^{-\beta_b(t-t_{vu})}, \quad \forall u \neq v
\end{split}
\raisetag{40pt}
\end{equation}
By comparing the form of the conditional intensity function for DLS \eqref{eq:dls_lambda} with that of our proposed LSH model \eqref{eq:lsh_lambda}, we identify 3 key differences, each addressing a concern regarding the DLS model:
\begin{enumerate}
\item The DLS utilizes reciprocal latent spaces $X^{(b)}$ to allow different rates of reciprocity between node pairs. 
This increase in flexibility of the model comes with a key drawback: the estimated latent positions for a node pair $(u,v)$ and kernel $b$ may result in the jump size $\alpha_2 e^{-||x_u^{(b)} - x_v^{(b)}||^2_2} > 1$, which leads to an unstable process. 
We were unable to simulate new networks from the DLS model fits to real networks due to the instability as we discuss in Section \ref{sec:generative_accuracy}. 
In contrast, we use just a single jump size $\alpha_2$ for all node pairs in our LSH model. 
While this may be less flexible, it does not lead to instability like the reciprocal latent space.

\item The DLS does not have a self excitation component, unlike our proposed LSH (second term in \eqref{eq:lsh_lambda}). 
The lack of self excitation prevents the DLS from modeling repeated edges from node $u$ to $v$ with no response from $v$ back to $u$. 
For example, this setting occurs frequently in militarized conflicts between countries, where one country repeatedly threatens or takes action against another country that does not retaliate.

\item The DLS does not have nodal effects parameters ($\delta_u$ and $\gamma_v$ in \eqref{eq:lsh_lambda}). 
This limits its ability to model nodes with different rates of sending or receiving events.
\end{enumerate}

Furthermore, a primary motivation of the latent space model is to embed the network into a single Euclidean space that can be easily visualized and interpreted. 
By using a single latent space, our proposed LSH is able to provide a much more interpretable model fit compared to DLS.

\section{Estimation Procedure}

Our model consists of mutually exciting bivariate Hawkes processes over all pairs of nodes. 
Using the likelihood theorem of \citet{daley2003introduction}, we can write the log-likelihood as
\begin{equation} \label{eq:3}
\log \mathcal{L} = \sum_{u\neq v} \left\{ \sum_{i=1}^k \log(\lambda_{uv}^*(t_i)) - \int_0^{t_k} \lambda_{uv}^* (t) dt \right\},
\end{equation}
where $k$ denotes the total number of events and $\lambda_{uv}^*(t)$ takes on the form in \eqref{eq:lsh_lambda}. 
We simplify the log-likelihood and improve the efficiency of the estimation by deriving a recursive form as in \cite{ozaki1979maximum}. More details and the full log-likelihood derivation for our LSH model are provided in Appendix \ref{sec:appendix_ll}, resulting in the simplified expression in \eqref{eq:5}. 

Latent space models typically assume that the probability of forming an edge between two nodes is inversely proportional to the distances between the node positions in the latent space.
Thus, the observation of an edge between two nodes typically pulls them closer together in the latent space.
The presence of the slope parameter $\theta_1$ in the baseline rate $\mu_{uv}$ for our LSH model \eqref{eq:baseline} allows us to either pull node pairs with events closer together by constraining $\theta_1 > 0$ or push them further apart by constraining $\theta_1 < 0$. 
Or, we could leave $\theta_1$ unconstrained---we find that this usually results in the estimate $\hat{\theta}_1 > 0$.

We use the L-BFGS-B algorithm \citep{byrd1995limited} to minimize the negative log-likelihood (NLL). 
The gradients of the log-likelihood can be carried out using the Autograd package \citep{maclaurin2015autograd} for automatic differentiation of standard Python functions. 
We consider also an alternating minimization approach that alternates between estimating the latent space and the model parameters, which we show in Appendix \ref{sec:alternating}. 
Our alternating minimization approach is partly inspired by the projected gradient method of \citet{ma2020universal}, which also alternates between estimating the latent space and the model parameters in a static
latent space model. 
We find that the alternating minimization approach generally converges more slowly than L-BFGS-B, so the results we present in this paper use L-BFGS-B.

\begin{figure*}[t]
\newcommand{\figwidth}{0.32\textwidth}
\newcommand{\figheight}{1.5in}
\centering 
    \hfill
    \begin{subfigure}[c]{\figwidth}
        \centering
        \includegraphics[height=\figheight]{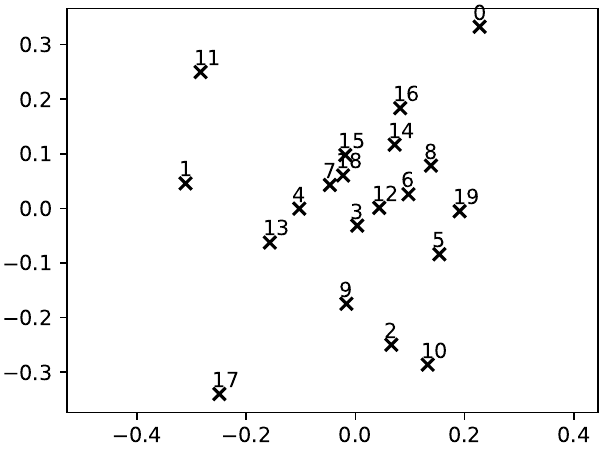}
        \caption{Actual latent space}
        \label{fig:sim1_actual}
    \end{subfigure}
    \hfill
    \begin{subfigure}[c]{\figwidth}
        \centering
        \includegraphics[height=\figheight]{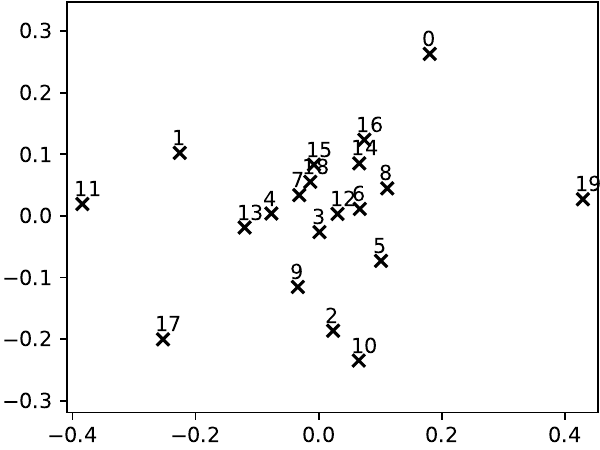}
        \caption{Estimated latent space}
        \label{fig:sim1_estimate}
    \end{subfigure}
    \hfill
    \begin{subfigure}[c]{0.34\textwidth}
        \centering
        \includegraphics[height=\figheight, clip=true, trim=5 10 25 25]{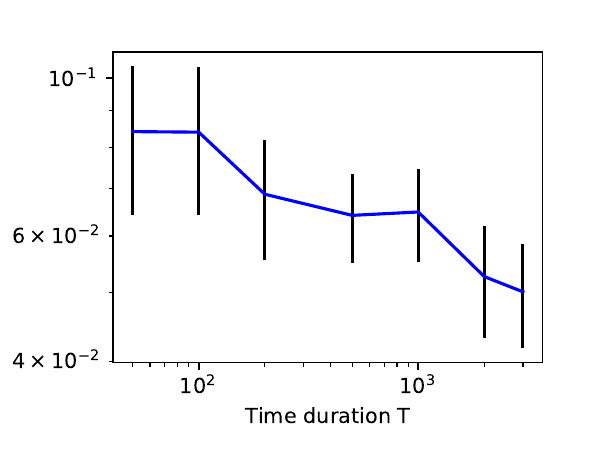}
        \caption{Latent positions estimation error}
        \label{fig:rmse_sim_z}
    \end{subfigure}
    \hfill
\caption{Comparison of \subref{fig:sim1_actual} actual latent space and \subref{fig:sim1_estimate} estimated latent space (with Procrustes transformation) on a $20$ node simulated network with duration $T=100$. 
The recovered latent node positions are close to the actual positions. 
\subref{fig:rmse_sim_z} The RMSE over $30$ simulated networks ($\pm \, 2$ standard errors) decreases as the duration $T$ increases.}
\label{fig:sim1}
\end{figure*}

We use a multidimensional scaling algorithm as an initialization for the latent space positions $Z$ as in the original latent space model proposed by \citet{hoff2002latent}. We set random values to initialize all other parameters $\Theta = (\alpha_1, \alpha_2, \theta_1, \theta_2, \delta, \gamma)$.

\section{Experiments}

In this section, we perform evaluation tasks for our proposed model on simulated networks and real networks\footnote{Python code to reproduce our results is available at \url{https://github.com/IdeasLabUT/Latent-Space-Hawkes}}.
We use a sum of $B = 3$ exponential kernels and utilize decays with time scales of an hour, a day, and a week, which is the same as \citet{yang2017decoupling} did in their DLS model. We also fix $C = (1/3, 1/3, 1/3)$ for simplicity\footnote{We also experimented with estimating $C$ but did not find much difference in the results.}.

\subsection{Simulated Networks}
\label{sec:sim_experiment}

We first test our L-BFGS-B estimation procedure on networks simulated from our latent space Hawkes (LSH) model. 
We simulate networks of $20$ nodes in a 2-D latent space using parameters $\theta_1 = 1$, $\theta_2 = -3.2$, $\alpha_1 =  0.01$, and $\alpha_2=0.02$.
Each dimension of the latent positions as well as sender and receiver effects for nodes are sampled independently from a standard Normal distribution: $z_{u}, \delta_u, \gamma_u \sim \mathcal{N}(0,1)$. 
We increase the time duration $T$ from $50$ to $3,000$ and evaluate the estimation accuracy for the latent positions and other parameters.
Additional details on the simulation process is provided in Appendix \ref{sec:sim_appendix}. 
A comparison of the actual and estimated latent positions for a simulated network is shown in Figure \ref{fig:sim1} along with the root mean squared error (RMSE) for estimated latent positions over 30 simulated networks. 
As expected, the error decreases for increasing time duration $T$. 
The error for the other parameters decreases also, as we show in Figure \ref{fig:sim1_supp} in Appendix \ref{sec:sim_appendix}. 
Thus, L-BFGS-B appears to accurately estimate latent positions and model parameters.

\subsection{Real Networks}

\begin{table}[t]
    \centering
    \caption{Summary statistics of real network datasets}
    \label{tab:dataStats}
    \begin{tabular}{cccc}
    \toprule
    Dataset  & Nodes    & Events & Time Duration \\
    \midrule
    Reality  & $65$     & $2,150$      & $8$ months \\
    Enron    & $155$    & $9,646$      & 15 months \\
    MID      & $145$    & $5,088$      & $23$ years \\
    FB-forum & $899$    & $33,720$    & $5.5$ months \\
    \bottomrule
    \end{tabular}
\end{table}

We perform experiments on several real network datasets: Reality Mining \citep{eagle2006reality}, Enron emails \citep{klimt2004enron}, Militarized Interstate Disputes (MID) \citep{palmer2021mid5}, and Facebook-forum \citep{nr}.
Summary statistics for the datasets are shown in Table \ref{tab:dataStats}, and  
additional details are provided in Appendix \ref{sec:supp_datasets}.  
Each dataset consists of a set of relational events, each denoted by a sender, a receiver, and a timestamp.

\paragraph{Baselines for Comparisons}
We compare against three other Hawkes process-based continuous-time network models. The dual latent space (DLS) model \citep{yang2017decoupling} is the most similar to ours, and we provide a detailed comparison of the DLS model with our proposed LSH model in Section \ref{sec:relation_DLS}. 
We also compare against two recently proposed Hawkes process-based block models: the community Hawkes independent pairs (CHIP) model \citep{arastuie2020chip} and the block Hawkes model (BHM) \citep{junuthula2019block}. 
Finally, we compare also against the non-generative continuous-time dynamic network embeddings (CTDNE) \citep{nguyen2018continuous} approach. 
Additional information on these models for comparison along with implementation details are provided in Appendix \ref{sec:supp_models}.

\subsubsection{Predictive Accuracy}
\label{sec:real_experiment}
We first evaluate the predictive ability of our proposed LSH model. 
We split each dataset into a training set containing the first $80\%$ of events and a test set containing the remaining $20\%$ of events. 
We estimate model parameters on the training set and evaluate prediction accuracy on the test set. 
We choose the number of latent dimensions $d$ (for LSH and DLS) and the number of blocks $K$ (for BHM and CHIP) that maximizes the log-likelihood evaluated on the test set. 

\paragraph{Test Log-likelihood}

\begin{table}[t]
  \caption{Evaluation metrics for predictive accuracy on real network datasets. 
  Bold entry denotes highest accuracy for each metric on a dataset. 
  Test log-lik.~shows the mean test set log-likelihood per event and the number of latent dimensions $d$ or blocks $K$ that maximize it. 
  The AUC column shows the mean (standard deviation) of the AUC across 100 time points for dynamic link prediction. 
  DLS does not scale to the FB-forum data.
  CTDNE is not generative so test log-likelihood is not applicable.
  }
  \label{tab:predictive_accuracy}
  \centering
  \setlength{\tabcolsep}{3pt}
  \begin{tabular}{cccccc}
    \toprule
    Dataset                   & Model  & Test log-lik.       & AUC  \\
    \midrule
    \multirow{5}{*}{Reality}  & LSH    & $\bm{-3.71}\,(d=4)$ & $0.945(0.028)$\\
                              & DLS    & $-5.64\,(d=300)$    & $0.940(0.034)$ \\
                              & BHM    & $-5.31\,(K=50)$     & $\bm{0.957(0.022)}$\\
                              & CHIP   & $-4.70\,(K=1)$      & $0.937(0.028)$\\
                              & CTDNE  &                     & $0.936(0.033)$\\
    \midrule
    \multirow{5}{*}{Enron}    & LSH    & $\bm{-4.87}\,(d=4)$ & $0.946(0.024)$ \\
                              & DLS    & $-5.29\,(d=100)$    & $\bm{0.947(0.017)}$\\
                              & BHM    & $-6.35\,(K=14)$     & $0.839(0.035)$\\
                              & CHIP   & $-5.34\,(K=4)$      & $0.895(0.053)$ \\
                              & CTDNE  &                     & $0.912(0.035)$\\

    \midrule
    \multirow{5}{*}{MID}     & LSH     & $\bm{-3.38}\,(d=3)$ & $\bm{0.988(0.018)}$\\
                             & DLS     & $-4.52\,(d=100)$    & $0.977(0.007)$\\
                             & BHM     & $-4.97\,(K=95)$     & $0.971(0.031)$ \\
                             & CHIP    & $-3.63\,(K=2)$      & $0.958(0.035)$\\
                             & CTDNE   &                     & $0.953(0.018)$\\
    \midrule
    \multirow{4}{*}{FB-forum}& LSH     & $\bm{-7.21}\,(d=8)$ & $\bm{0.932(0.009)}$  \\
                             & BHM     & $-11.16\,(K=57)$    & $0.839(0.017)$ \\
                             & CHIP    & $-7.65\,(K=2)$      & $0.919(0.011)$ \\
                             & CTDNE   &                     & $0.788(0.028)$\\
    \bottomrule
  \end{tabular}
\end{table}

We use the mean log-likelihood per event on the test set, also used by \citet{Dubois2013} and \citet{arastuie2020chip}, as an evaluation metric for the model's predictive ability on future data. 
As shown in Table \ref{tab:predictive_accuracy}, our Latent Space Hawkes (LSH) significantly outperforms the other models on all datasets. 
The test log-likelihood is maximized for the LSH at relatively small latent dimensions $d$ compared to the DLS model. 
The low-dimensional latent representation using a single latent space makes the LSH fit more interpretable than the high-dimensional DLS representation using multiple latent spaces. 
Furthermore, these results suggest that the addition of nodal effects and self excitation in the LSH significantly affects the predictive ability compared to DLS.

\paragraph{Dynamic Link Prediction}
We further explore the performance of the learned model in a dynamic link prediction task. 
We use the same experiment set-up as \citet{yang2017decoupling}. 
We randomly sample 100 time points $t_i$ during the test period. We then compute the probability of a link appearing between each pair of nodes in the $[t_i, t_i+\delta$] time window. We set $\delta$ to be two weeks for the Reality, Enron, and FB-forum datasets and two months for the MID data, which takes place over a longer period of time. For each of these $\delta$ intervals, we obtain the Receiver Operating Characteristics (ROC) curve and compute the Area Under the Curve (AUC) measured across all pairs of nodes according to the predicted probabilities given by the model. 

The mean AUC values are shown in Table \ref{tab:predictive_accuracy} with the value inside the parentheses indicating the standard deviation over these 100 time intervals. 
The ROC curves and box plots for the corresponding AUC values are presented in Appendix \ref{sec:supp_dynamic_lp}.
Our proposed LSH model is competitive at the dynamic link prediction task, achieving highest mean AUC on FB-forum and MID and second highest on Reality and Enron.

\subsubsection{Generative Accuracy}
\label{sec:generative_accuracy}

To evaluate generative accuracy of our proposed LSH model, we simulate networks with our fitted parameters and perform posterior predictive checks (PPCs) using network statistics such as  reciprocity and transitivity.  
While our LSH model has no issues simulating networks, the DLS is problematic due to its model formulation. 
The jump size for reciprocal excitation depends on distances between nodes in a reciprocal latent space and is further scaled by the parameter $\alpha_2$ in \eqref{eq:dls_lambda}. 
Since the maximum jump size is not constrained, this results in some node pairs having unstable Hawkes processes so that the simulation does not terminate. 
To enable us to make comparisons with the DLS model, we stabilize it by fixing the scaling parameter for the jump size $\alpha_2 = 1$. 

We simulate 15 networks from the fitted model on each real dataset, with the exception of DLS, which does not scale to the FB-forum data. 
We then perform PPCs on the number of events generated, average run length, and 4 static network statistics: transitivity (global clustering coefficient), reciprocity, average local clustering coefficient (LCC), and average degree. 
The run length is the number of consecutive events in the same direction, e.g.~in the sequence $(u,v), (v,u), (v,u), (v,u), (v,u), (u,v)$, the run length for $(v,u)$ is 4 because it appears 4 times consecutively before the reciprocal event $(u,v)$ appears.

\begin{table}[t]
\caption{Comparison of generative accuracy between models using mean statistic over 15 simulated networks.
Bold entry denotes the simulated statistic closest to the actual statistic. 
While both LSH and DLS can replicate the static network statistics from the actual networks, DLS generates way too many events compared to the actual networks.}
\label{tab:generative_accuracy}
\centering
\setlength{\tabcolsep}{5pt}
    \begin{tabular}{ccccc}
    \toprule
    Dataset                  & Statistic       & Actual     & LSH             & DLS       \\ 
    \midrule
    \multirow{6}{*}{Reality} & \# of events    & $2,\!148$  & $\bm{2,\!190}$  & $9,\!493$   \\
                             & Avg.~run length & $2.49$     & $\bm{2.62}$     & $1.91$      \\
                             & Transitivity    & $0.29$     & $0.34$          & $\bm{0.32}$      \\
                             & Reciprocity     & $0.80$     & $\bm{0.86}$     & $0.52$      \\
                             & Avg.~LCC        & $0.25$     & $0.19$          & $\bm{0.21}$      \\
                             & Avg.~degree     & $4.86$     & $\bm{4.45}$     & $7.50$      \\
    \midrule
    \multirow{6}{*}{Enron}   & \# of events    & $9,\!646$  & $\bm{11,\!010}$ & $675,\!621$ \\
                             & Avg.~run length & $2.44$     & $\bm{2.63}$     & $1.87$      \\
                             & Transitivity    & $0.31$     & $0.39$          & $\bm{0.30}$      \\
                             & Reciprocity     & $0.65$     & $\bm{0.65}$     & $\bm{0.65}$      \\
                             & Avg.~LCC        & $0.40$     & $0.51$          & $\bm{0.36}$      \\
                             & Avg.~degree     & $18.46$    & $25.86$         & $\bm{18.43}$     \\ 
    \midrule
    \multirow{6}{*}{MID}     & \# of events    & $5,\!088$  & $\bm{3,\!996}$  & $412,\!890$ \\
                             & Avg.~run length & $2.88$     & $\bm{2.71}$     & $1.89$      \\
                             & Transitivity    & $0.13$     & $0.24$          & $\bm{0.20}$      \\
                             & Reciprocity     & $0.64$     & $\bm{0.57}$     & $0.52$      \\
                             & Avg.~coef       & $0.25$     & $\bm{0.29}$     & $\bm{0.29}$      \\
                             & Avg.~degree     & $6.80$     & $\bm{7.05}$     & $9.57$      \\ 
    \bottomrule
    \end{tabular}
\end{table}
A comparison between the actual statistics and mean simulated statistics is shown in Table \ref{tab:generative_accuracy}. We compare LSH and DLS since they are both based on the latent space model. The DLS model generates significantly more events than exist in the actual network, ranging from roughly a 4x increase (Reality) to an 80x increase (MID). 
We believe that this is due to the reciprocal latent space used in the DLS model. 
Even though we stabilized the model by setting $\alpha_2 = 1$, some nodes are likely still extremely close in the reciprocal latent space, causing too many events to be generated. 

We also find that the lack of self-excitation in DLS prevents it from replicating the run length of directed event sequences. Since DLS only has reciprocal excitation, its generated networks have the average run length of about 2 regardless of the average run length in the actual network. 
On the other hand, the DLS model performs quite well at replicating the static network statistics, and in many cases, even better than our proposed LSH. 
We believe that this is partially due to the much higher latent dimension $d$ that maximizes the test log-likelihood for DLS. 
The LSH could potentially achieve better generative accuracy using higher $d$ as well.
Additional results on generative accuracy, including plots comparing the actual statistics with the distribution of the simulated statistics, are provided in Appendix \ref{sec:supp_ppc}.

\section{Case Study}

\begin{figure*}[p]
\centering 
\newcommand{\figwidth}{4.65in}
    \begin{subfigure}[c]{\figwidth}
        \centering
        \hfill
        \includegraphics[width=\figwidth]{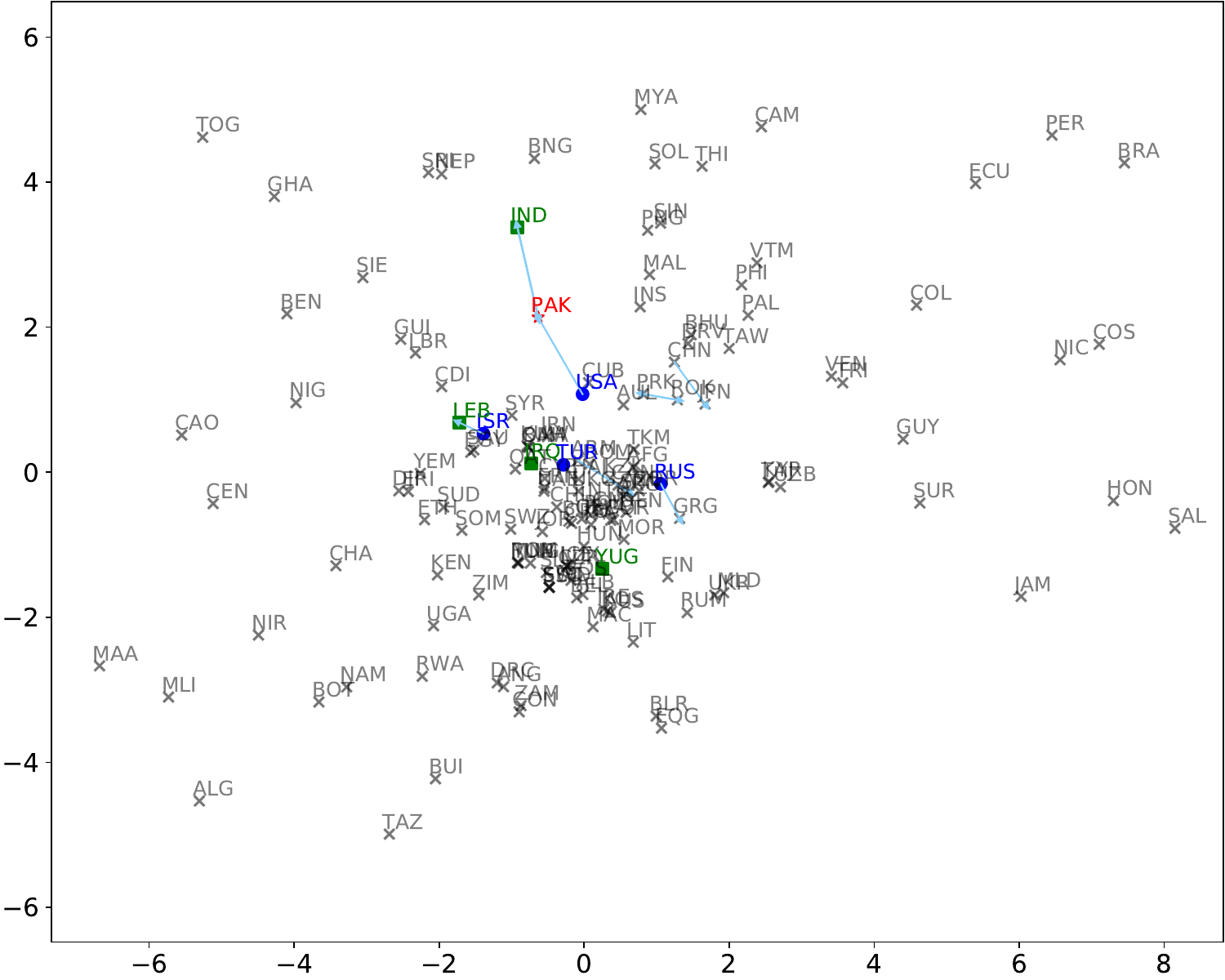}
        \caption{Estimated latent positions from model with positive slope}
        \label{fig:mid_latent_pos}
    \end{subfigure}
    \\[12pt]
    \begin{subfigure}[c]{\figwidth}
        \centering
        \hfill
        \includegraphics[width=\figwidth]{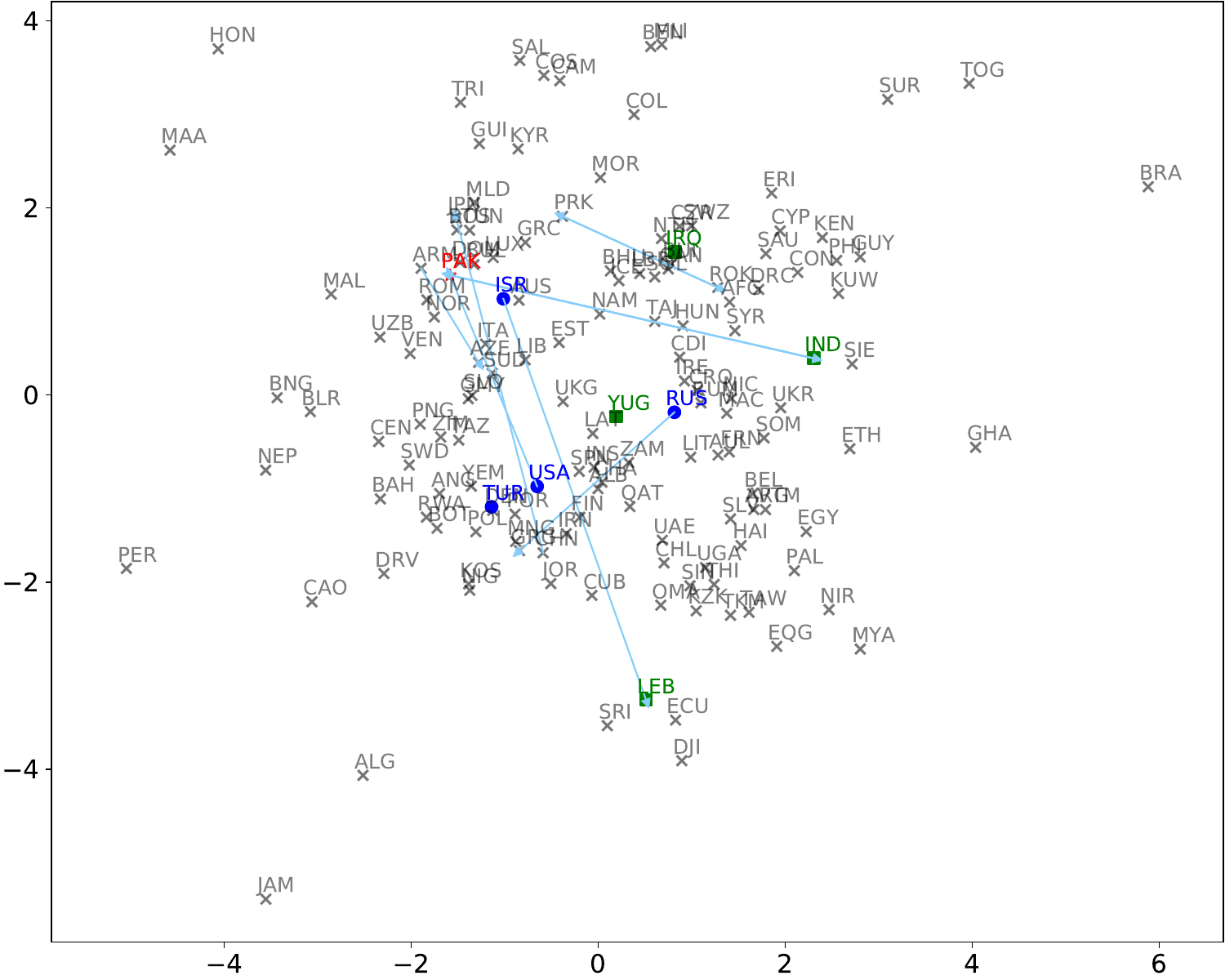}
        \caption{Estimated latent positions from model with negative slope}
        \label{fig:mid_latent_neg}
    \end{subfigure}
\caption{2-D latent space plots for LSH model fit to MID data. 
Edges are shown for the 10 most frequently occurring incidents. 
The most active countries that initiate and receive the 5 most incidents are shown in blue and green, respectively. 
Pakistan (PAK) is among the top 5 initiators and receivers and is shown in red.
\subref{fig:mid_latent_pos} The model with positive slope places countries with lots of conflicts close together. 
The most active countries tend to appear centrally in this latent space. 
A zoomed in version of the center of the latent space is shown in Figure \ref{fig:mid_latent_zoomed} in Appendix \ref{sec:supp_case_study}. 
\subref{fig:mid_latent_neg} The model with negative slope places countries with lots of conflicts far apart. 
The most active countries tend to appear on the periphery of this latent space.}
\label{fig:mid_latent}
\end{figure*}

\begin{figure*}[t]
\centering 
\newcommand{\figwidth}{4.65in}
        \includegraphics[width=\figwidth]{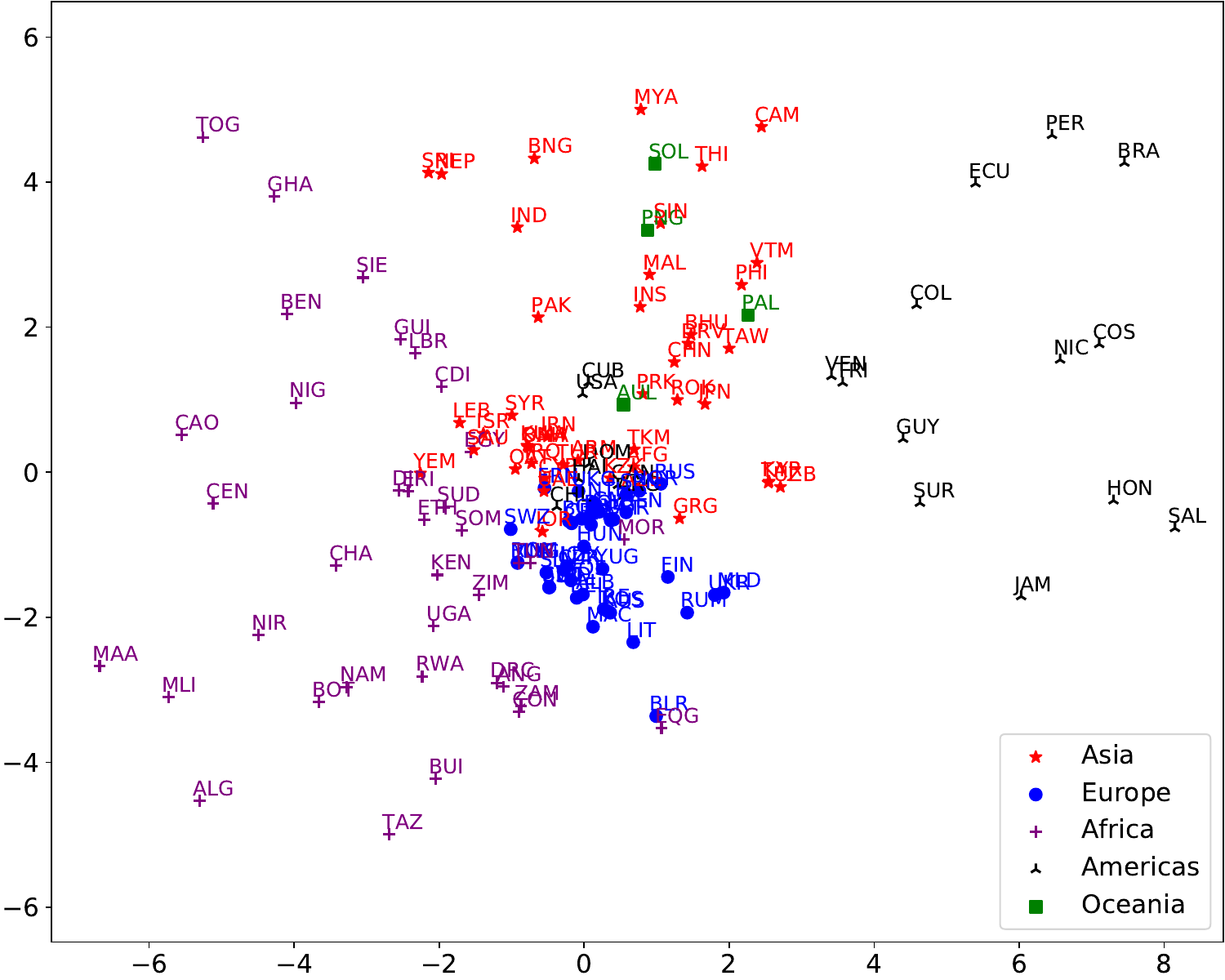}
\caption{2-D latent space plot for MID data with positive slope and countries colored by continent. A zoomed in version of the center of the latent space is shown in Figure \ref{fig:mid_latent_cont_zoomed} in Appendix \ref{sec:supp_case_study}.}
\label{fig:mid_latent_continents}
\end{figure*}

We now present a case study demonstrating our proposed LSH model being used for exploratory analysis on a real continuous-time network: the Militarized Interstate Disputes (MID) incident network. 
Timestamped edges in this network correspond to individual incidents within disputes between countries. 
Incidents include threats, displays, and uses of force initiated by one country towards another. 

Incidents in the MID network are indicative of negative relationships between countries. 
As a result, one might expect the network to be disassortative. 
On the other hand, incidents frequently occur between countries that are geographically close, particularly if they share a boundary, which suggests that the network may also have an assortative structure. 
Thus, we conduct exploratory analysis of this network using two different parameterizations of our model. 
We fix the latent dimension to be $d=2$ in both models so that we can visualize the latent positions of the countries. 

We first consider a \emph{positive slope} model by constraining $\theta_1 > 0$ in \eqref{eq:baseline} so that two countries with lots of incidents between them are pulled closer together in the latent space, as is typically the case for assortative networks. 
In this parameterization, countries that engage in lots of incidents are likely to appear centrally in the latent space. 
We next consider a \emph{negative slope} model by constraining $\theta_1 < 0$ in \eqref{eq:baseline} so that two countries with lots of incidents between them are pushed further apart in the latent space. 
Under this parameterization, countries that engage in lots of incidents are likely to appear on the periphery of the latent space.

\paragraph{Findings and Discussion}
We show the 2-D latent space plot with both positive and negative slope terms in Figure \ref{fig:mid_latent}. 
We first consider the latent positions from the positive slope model. 
Notice that the most active nodes tend to appear centrally, and the node pairs with the most frequent incidents tend to be placed close together. 
For example, Israel (ISR) and Lebanon (LEB) have latent positions very close together, which makes sense given that they have the most incidents in the data set: 588 total incidents.
Additionally, countries that are geographically close do mostly appear close together in the latent space. 
This can be seen from Figure \ref{fig:mid_latent_continents}, where nodes are colored by continent. 
The estimated parameters are $\theta_1 = 1.2, \theta_2 = -9.3, \alpha_1 = 0.77, \alpha_2 = 0.13$. 
The high value for $\alpha_1$ compared to $\alpha_2$ indicates the importance of self excitation in addition to reciprocal excitation.

Next, we consider the negative slope model. 
From examining the latent positions, we find that most active nodes tend to appear on the periphery of the latent space, which is reasonable because the model attempts to push nodes with many incidents far apart. 
For example, Israel and Lebanon are on opposite sides of the latent space.  
The estimated parameters for this model are $\theta_1 = -0.008, \theta_2 = -1.24, \alpha_1 = 0.83, \alpha_2 = 0.15$. 
While the parameters used for modeling the baseline intensity have changed significantly, the $\alpha$ parameters modeling self and reciprocal excitation are very similar to the positive slope model.

Additional results are presented in Appendix \ref{sec:supp_case_study}. 
We note that this case study is intended to be exploratory rather than confirmatory. 
We caution readers from jumping to conclusions about countries from our results.

\section{Conclusion}
We proposed the latent space Hawkes (LSH) model for continuous-time networks of relational events, 
which models interactions between each pair of nodes as realizations from a mutually exciting Hawkes processes whose intensity functions include a baseline rate along with both self and reciprocal excitation terms. 
The LSH model makes use of a single latent space along with sender and receiver effects to provide a more interpretable fit while remaining competitive in accuracy compared to the dual latent space (DLS) model. 
We performed an exploratory analysis of militarized disputes between countries using the LSH, where the latent space was quite informative of the dispute network structure. 
We also found that self excitation was stronger than reciprocal excitation in this network, demonstrating the importance of self excitation, which is not present in the DLS model. 
We hope this paper inspires future work combining continuous latent space representations with TPPs, which have not gotten as much attention as block model-based TPPs.

\paragraph{Limitations}
While our proposed model shows superior empirical performance and interpretability, there are also several limitations. 
We use a single reciprocal jump size $\alpha_2$ for all node pairs, which results in a less flexible model compared to the DLS, but it is
more stable. 
While our estimation procedure scales to networks with about $1,000$ nodes, it does not scale to extremely
large networks with $>10,000$ nodes, unlike the the CHIP \citep{arastuie2020chip} and MULCH \citep{soliman2022multivariate} latent block models. 
Additionally, the latent positions of nodes in our LSH model are fixed over time, just like in the DLS. If there are significant changes in the
network structure over time, a more flexible model that allows latent positions to change over time, such as the model of \citet{rastelli2021continuous}, may be a
better fit.
Finally, one could model more complex dependencies among the nodes that goes beyond self and reciprocal
excitation using a multivariate Hawkes process, as in the MULCH latent block model \citep{soliman2022multivariate}, instead of a bivariate Hawkes process.

\begin{contributions} 

Z.~Huang, S.~Paul, and K.~S.~Xu contributed to the model and algorithm development. 
Z.~Huang and H.~Soliman wrote the code and developed the experiments. 
All authors contributed to writing the paper.
\end{contributions}

\begin{acknowledgements} 

This material is based upon work supported by the National Science Foundation grants IIS-1755824, DMS-1830412, IIS-2047955, and DMS-1830547. 

\end{acknowledgements}

\newpage
\bibliography{reference}

\newpage
\appendix
\onecolumn

\newpage
\section{Additional Model and Estimation Details}

\subsection{Proof of Theorem \ref{thm:identifiability}}
\label{sec:supp_proof}

\begin{proof}
Suppose two sets of parameters $\theta_2, Z,\gamma,\delta$ and $\theta_2', Z',\gamma',\delta'$ lead to the same $\log (\mu)$:
\begin{equation*}
\theta_2 J_n + ZZ^T + \tilde{\delta}1_n^T + 1_n \tilde{\gamma}^T = \theta_2' J_n + Z'Z'^T + \tilde{\delta}'1_n^T + 1_n \tilde{\gamma}'^T.
\end{equation*}
Left multiplying both sides by $1_n^T$ and noting that (i) $1_n^T Z = 1_n^T H Z = 1_n^T (I_n - \frac{1}{n}1_n 1_n^T)Z = 0$ by assumption 1 and (ii) $1_n^T \tilde{\delta}=0$ by assumption 2,  we get
\begin{align}
& 1_n^T\theta_2 J_n  + 1_n^T 1_n \tilde{\gamma}^T = 1_n^T \theta_2' J_n  + 1_n^T 1_n \tilde{\gamma}'^T \nonumber\\
& \Rightarrow n (\theta_2 - \theta_2')1_n^T  + n(\tilde{\gamma}^T - \tilde{\gamma}'^T) = 0
\label{eq:identify}
\end{align}
Now, right multiplying by $1_n$ we get 
\begin{equation*}
n^2 (\theta_2 - \theta_2') =0
\end{equation*}
because $\gamma^T 1_n = 0$ and $\tilde{\gamma}^T 1_n = 0$ by our identifiability constraints.
This implies that $\theta_2 = \theta_2'$. With this, using \ref{eq:identify}, we have
\begin{equation*}
n(\tilde{\gamma}^T - \tilde{\gamma}'^T) = 0 \Rightarrow \tilde{\gamma} =\tilde{\gamma}'.
\end{equation*}
Finally, right multiplying by $1_n$ from the beginning and then using the above result $\theta_2 = \theta_2'$ we have
\begin{align*}
& \theta_2 J_n1_n  + \tilde{\delta} 1_n^T 1_n  =  \theta_2' J_n 1_n +  \tilde{\delta}' 1_n^T1_n \\
& \Rightarrow  n(\tilde{\delta} - \tilde{\delta}') = 0, \quad \Rightarrow \tilde{\delta} = \tilde{\delta}'.
\end{align*}
In light of the above results we then conclude
\[
ZZ^T = Z'Z'^T. 
\]
which means $Z=Z'O$ where $O_{d \times d}$ is an orthogonal matrix such that $OO^T =I$.
Furthermore, from the results of \citet{ozaki1979maximum}, the baseline intensity $\mu_{uv}$ and jump size parameters $\alpha_1, \alpha_2$ are
identified, i.e.~if two sets of $\mu_{uv}, \alpha_1, \alpha_2$ lead to the same probability density function (log-likelihood), then the two
sets must be identical.
\end{proof}

\subsection{Full log-likelihood expression}
\label{sec:appendix_ll}

Let $\Lambda_{uv}(t_k^{uv}) = \int_0^{t_k} \lambda_{uv}^* (t) \,dt$. We can break the time interval $[0,t_k]$ to $[0,t_1],[t_1, t_2], \ldots , [t_{k-1}, t_k]$, resulting in
\begin{equation}
\begin{split}
\Lambda_{uv}(t_k^{uv}) & = \int_0^{t_k^{uv}} \lambda_{uv}^* (t) \,dt \\
&= \int_0^{t_1^{uv}} \lambda_{uv}^*(t)\,dt + \sum^{k-1}_{i=1} \int^{t_{i+1}^{uv}}_{t_i^{uv}}  \lambda^*_{uv}(t)\,dt  \\
                  & = \int_0^{t_1^{uv}}\mu_{uv} \,dt + \sum^{k-1}_{i=1} \int^{t_{i+1}^{uv}}_{t_i^{uv}} \left[\mu_{uv} +  \sum_{j:t^{uv}_j < t_i^{uv}}\sum_b^B C_b\alpha_1\beta_b e^{-\beta_b(t-t^{uv}_j)} + \sum_{j:t^{vu}_j < t^{uv}_i}\sum_b^B C_b\alpha_2\beta_b e^{-\beta_b(t-t^{vu}_j)}\right]dt \\
                  & = \mu_{uv}t_k^{uv} + \sum^{k-1}_{i=1} \int^{t_{i+1}^{uv}}_{t_i^{uv}}\sum_b^B \left[ \sum_{j:t^{uv}_j < t_i^{uv}} C_b\alpha_1\beta_b e^{-\beta_b(t-t^{uv}_j)} + \sum_{j:t^{vu}_j < t^{uv}_i} C_b\alpha_2\beta_b e^{-\beta_b(t-t^{vu}_j)}\right]dt \\
                  & = \mu_{uv}t_k^{uv} + \sum^{k-1}_{i=1}\sum_{j:t^{uv}_j < t_i^{uv}}\sum_b^B \int^{t_{i+1}^{uv}}_{t_i^{uv}} \left[  C_b\alpha_1\beta_b e^{-\beta_b(t-t^{uv}_j)}\right] dt +    \sum^{k-1}_{i=1}\sum_{j:t^{vu}_j < t^{uv}_i}\sum_b^B \int^{t_{i+1}^{uv}}_{t_i^{uv}}\left[ C_b\alpha_2\beta_b e^{-\beta_b (t-t^{vu}_j)}\right]dt \\
                  & = \mu_{uv}t_k^{uv} - \sum^{k-1}_{i=1}\sum_b^B\sum_{j:t^{uv}_j < t_i^{uv}}{C_b\alpha_1} \left[ e^{-\beta_b(t_{i+1}^{uv}-t^{uv}_j)} - e^{-\beta_b(t_i^{uv}-t^{uv}_j)}\right] \\
                  &\qquad\qquad\qquad - \sum^{k-1}_{i=1}\sum_b^B\sum_{j:t^{vu}_j < t^{uv}_i}{C_b\alpha_2} \left[  e^{-\beta_b(t_{i+1}^{uv}-t^{vu}_j)} - e^{-\beta_b(t_i^{uv}-t^{vu}_j)}\right]
\end{split}
\end{equation}

By extending the summation, many of terms can cancel out, and then we can simplify $\Lambda_{uv}(t_k^{uv})$ as follows:
\begin{equation} \label{eq:4}
\Lambda_{uv}(t_k^{uv}) = \mu _{uv} t_k^{uv} - \sum_b^B C_b\alpha_1 \sum_{j:t^{uv}_j<t_k^{uv}} \left[ e^{-\beta_b(t_k^{uv} - t^{uv}_j)} -1 \right] - \sum_b^B C_b\alpha_2 \sum_{j:t^{vu}_j<t_k^{uv}} \left[ e^{-\beta_b(t_k^{uv} - t^{vu}_j)} -1 \right]. 
\end{equation}
Substituting \eqref{eq:4} into \eqref{eq:3} gives the following simplified expression for the log-likelihood.
\begin{equation} \label{eq:5}
\begin{split}
l = \sum_{u\neq v} \Bigg\{ & \sum_{i=1}^k \log\left[\mu_{uv} +  \sum_{j:t^{uv}_j < t_i^{uv}}\sum_b^B  C_b\alpha_1\beta_b e^{-\beta_b(t_i^{uv}-t^{uv}_j)} + \sum_{j:t^{vu}_j < t^{uv}_i} \sum_b^B C_b\alpha_2\beta_b e^{-\beta_b(t^{uv}_i-t^{vu}_j)}\right] \\
    & -\mu _{uv} t_k^{uv} + \sum_b^B C_b\alpha_1 \sum_{j:t^{uv}_j<t_k^{uv}} \left[ e^{-\beta_b(t_k^{uv} - t^{uv}_j)} -1 \right] - \sum_b^B C_b\alpha_2 \sum_{j:t^{vu}_j<t_k^{uv}} \left[ e^{-\beta_b(t_k^{uv} - t^{vu}_j)} -1 \right]   \Bigg\}.
\end{split}
\end{equation}
We use the negative of the log-likelihood expression in \eqref{eq:5} as the objective function for the L-BFGS-B minimization.

\subsection{Alternating Minimization}
\label{sec:alternating}

An alternative approach to obtaining the maximum likelihood estimate (MLE) is to partition our parameters into two sets: the latent node positions $Z$ and all other parameters $\Theta = (\alpha_1, \alpha_2, \theta_1, \theta_2, \delta, \gamma)$. 
We propose an alternating minimization approach to obtain the MLE. 
It alternates between optimizing the NLL over the estimated latent positions $\hat{Z}$ while holding all other parameters fixed and optimizing all other parameters $\hat{\Theta}$ while holding the latent positions fixed. 

We run each minimization over a fixed number of iterations and then switch to the other minimization. 
We experiment with different values for the number of steps $(s_{\Theta}, s_Z)$ denoting the number of iterations to run the optimization over $\Theta$ and $Z$, respectively. 
We find that there is not much difference in the performance for different numbers of steps for the directions in alternate minimization.  Taking 2 steps in each directions, i.e.~$s_{\Theta} = 2, s_Z = 2$, seems to work well.
Pseudocode for our alternating minimization estimation procedure is shown in Algorithm \ref{alg:alt_min}.

\begin{algorithm}[t]
   \caption{Alternating minimization estimation algorithm}
   \label{alg:alt_min}
\begin{algorithmic}[1]
   \STATE {\bfseries Input:} Relational event triplets $(u, v, t)$, latent dimension $d$, kernel decays $\beta$, kernel scale parameters $C$
   \STATE Initialize $ \hat{Z}$ using multidimensional scaling
   \STATE Initialize $ \hat{\Theta}$ randomly
   \REPEAT
   \STATE $\hat{\Theta}\leftarrow$ $s_\theta$ steps of L-BFGS-B over $\hat{\Theta}$ on log-likelihood \eqref{eq:5} while keeping $\hat{Z}$ fixed\;
   \STATE $\hat{Z}\leftarrow$ $s_Z$ steps of L-BFGS-B over $\hat{Z}$ on log-likelihood \eqref{eq:5} while keeping $\hat{\Theta}$ fixed\;
   \UNTIL{log-likelihood \eqref{eq:5} converges}
   \STATE {\bfseries Return:} Estimated model parameters $(\hat{Z}, \hat{\Theta})$
\end{algorithmic}
\end{algorithm}

\section{Additional Experiment Results}

\subsection{Simulated Networks}
\label{sec:sim_appendix}

The generative process for the simulated networks is shown in Algorithm \ref{alg:sim_exp}.
As discussed in Theorem \ref{thm:identifiability}, the latent positions $Z$ are only identifiable up to a rotation. 
Furthermore, the slope parameter $\theta_1$ is not identifiable, so we absorb it into the latent positions by setting
\begin{equation*}
\hat{Z} \leftarrow \hat{Z} \sqrt{|\hat{\theta}_1}| \qquad \text{and} \qquad \hat{\theta}_1 \leftarrow \frac{\hat{\theta}_1}  {\sqrt{|\hat{\theta}_1|}}. 
\end{equation*}
Then, to compare the estimated and actual latent positions, we apply a Procrustes transform \citep{gower1975generalized} to the estimated latent positions $\hat{Z}$. 

\begin{algorithm}[t]
   \caption{Generative procedure for Latent space Hawkes network}
   \label{alg:sim_exp}
\begin{algorithmic}[1]
   \STATE {\bfseries Input:} Number of nodes $n$, time duration $T$, latent dimension $d$, kernel decays $\beta$, kernel scaling parameters $C$, model parameters $(\theta_1,\theta_2, \alpha_1, \alpha_2)$
   \STATE Sample latent positions $Z$ and sender and receiver effects $\delta$, $\gamma$ from Normal distributions: $Z\leftarrow \mathcal{N}(0,\sigma_z^2I_{nd}), \delta \leftarrow \mathcal{N}(0,\sigma_\delta^2I_n), \gamma \leftarrow \mathcal{N}(0,\sigma_\gamma^2I_n)$
   \STATE Set $Z \leftarrow Z \sqrt{|\theta_1}|, \theta_1 \leftarrow \theta_1 / \sqrt{|\theta_1|}$ to remove identifiability issues
   \FORALL{node pairs $u \neq v$}
   \STATE $ \mu_{uv} \leftarrow e^{-\theta_1||z_u - z_v||^2_2 + \theta_2 + \delta_u
    +\gamma_v}$
   \STATE $\mathcal{H}(u,v) \leftarrow$ Ogata's thinning algorithm ($\mu_{uv}, \alpha_1, \alpha_2, T, \beta, C$)\;
   \ENDFOR

   \STATE {\bfseries Return:} Simulated network $\mathcal{H}$ containing events $\mathcal{H}(u,v)$ for all $u \neq v$
\end{algorithmic}
\end{algorithm}

\begin{figure*}[p]
\centering 
    \begin{subfigure}[c]{2.5in}
        \newcommand{\figwidth}{2.5in}
        \centering
        \hfill
        \includegraphics[width=\figwidth]{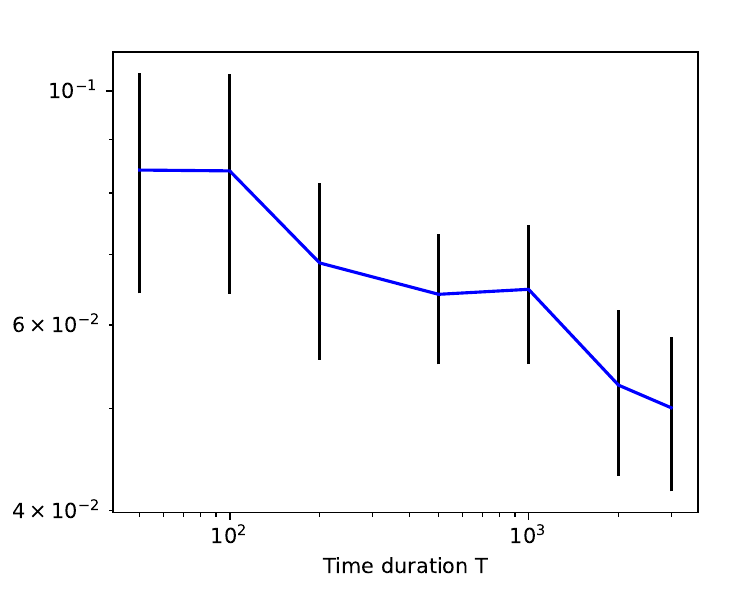}
        \caption{latent positions $z$}
    \end{subfigure}
    \begin{subfigure}[c]{2.5in}
        \newcommand{\figwidth}{2.5in}
        \centering
        \hfill
        \includegraphics[width=\figwidth]{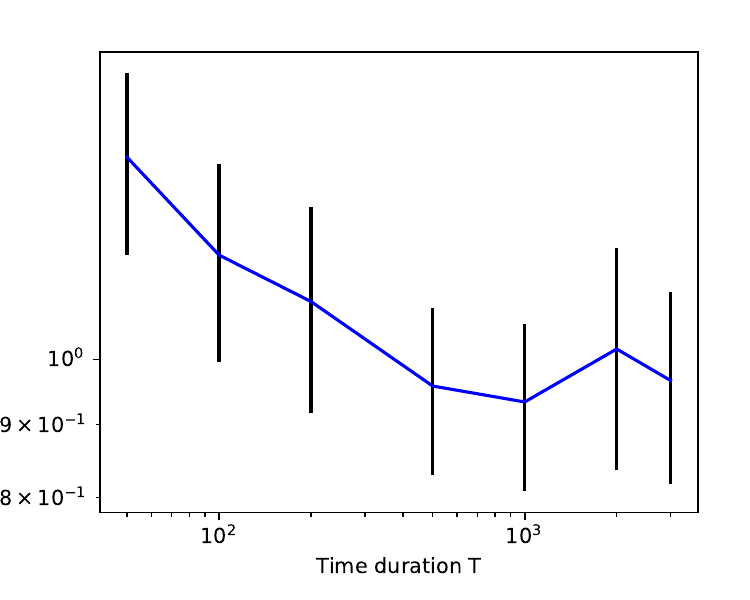}
        \caption{sender effects $\delta$}
    \end{subfigure}
        \begin{subfigure}[c]{2.5in}
        \newcommand{\figwidth}{2.5in}
        \centering
        \hfill
        \includegraphics[width=\figwidth]{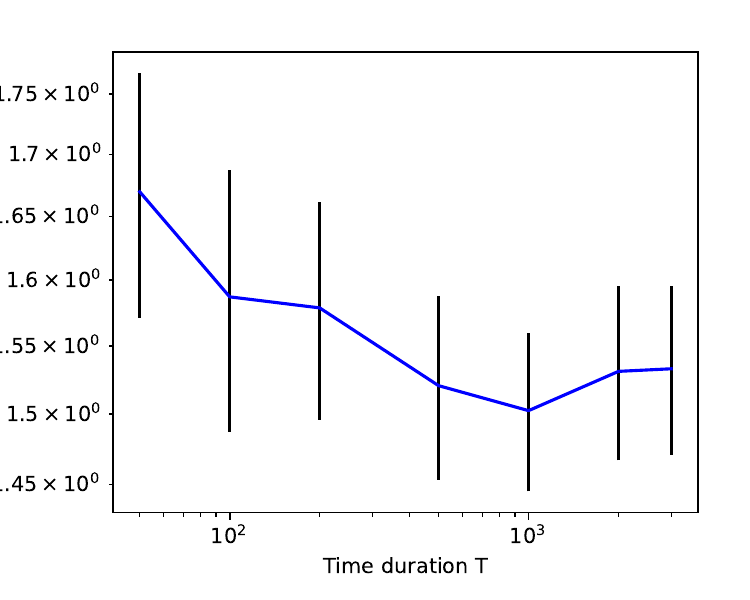}
        \caption{receiver effects $\gamma$}
    \end{subfigure}
    \begin{subfigure}[c]{2.5in}
        \newcommand{\figwidth}{2.5in}
        \centering
        \hfill
        \includegraphics[width=\figwidth]{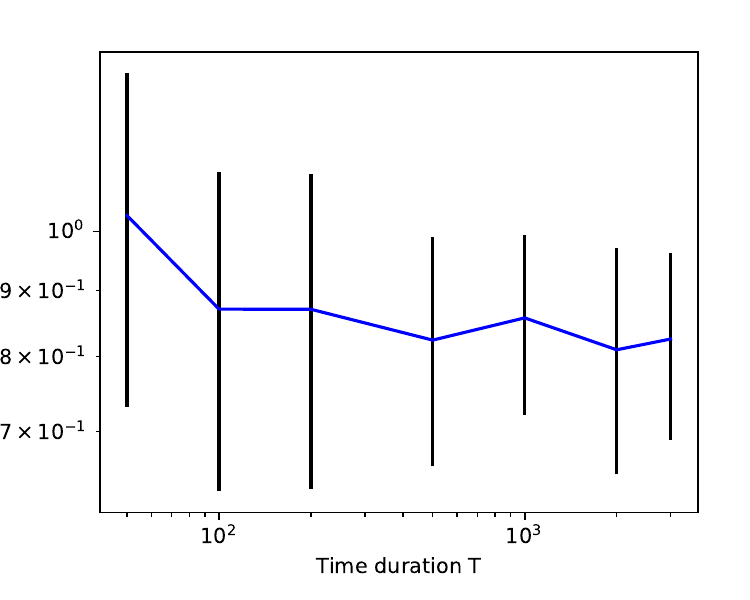}
        \caption{intercept parameter $\theta_2$}
    \end{subfigure}
    \begin{subfigure}[c]{2.5in}
        \newcommand{\figwidth}{2.5in}
        \centering
        \hfill
        \includegraphics[width=\figwidth]{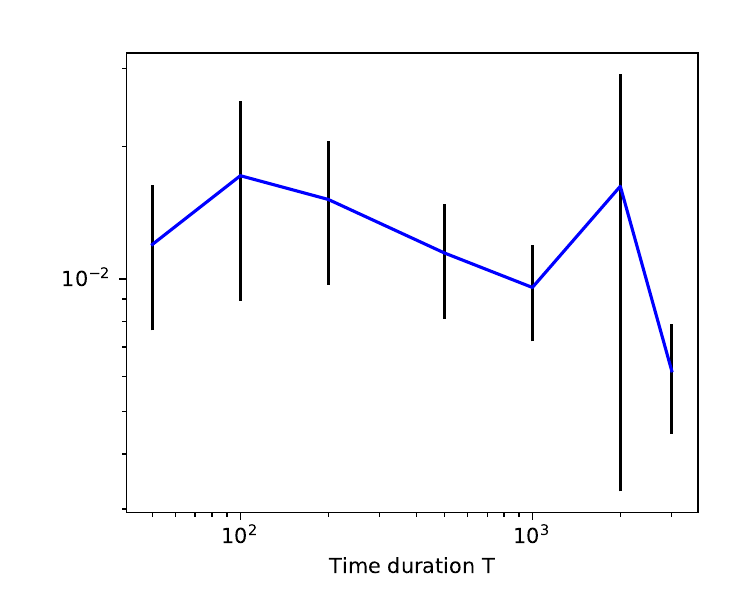}
        \caption{jump size for self-exciting term $\alpha_1$}
    \end{subfigure}
    \begin{subfigure}[c]{2.5in}
        \newcommand{\figwidth}{2.5in}
        \centering
        \hfill
        \includegraphics[width=\figwidth]{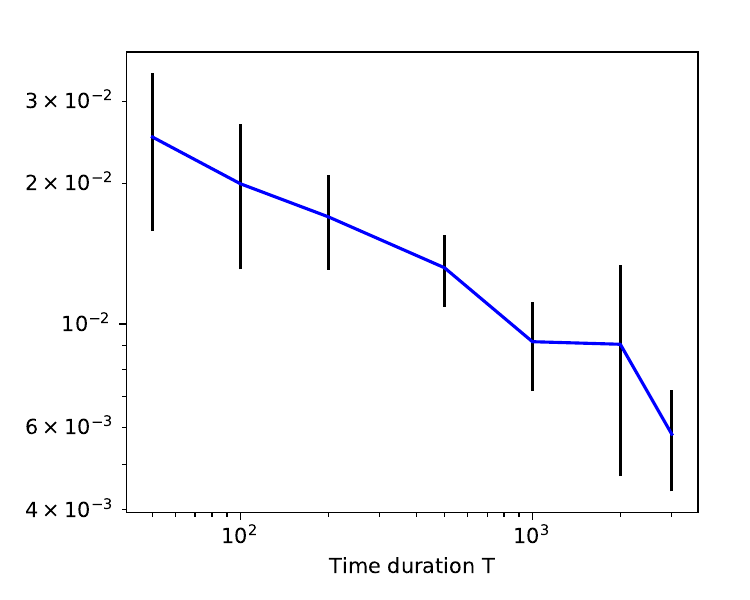}
        \caption{jump size for reciprocal term $\alpha_2$}
    \end{subfigure}
    \begin{subfigure}[c]{2.5in}
        \newcommand{\figwidth}{2.5in}
        \centering
        \hfill
        \includegraphics[width=\figwidth]{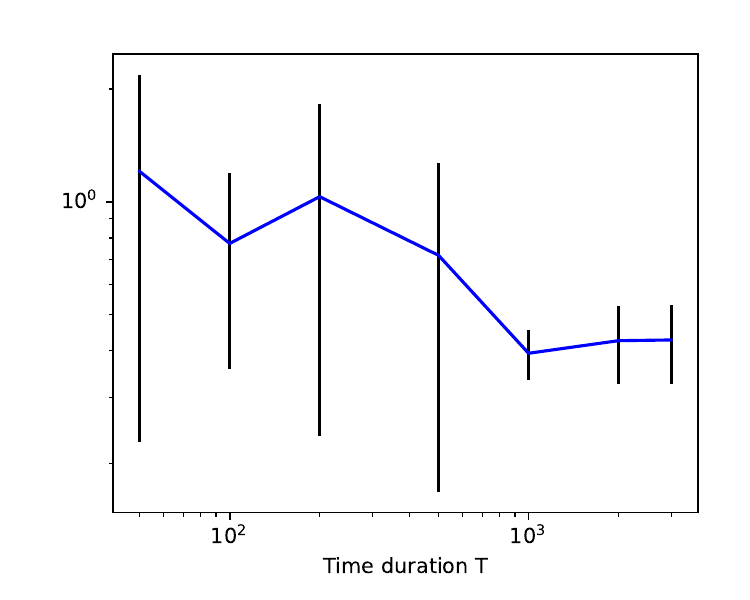}
        \caption{baseline intensity $\mu$}
    \end{subfigure}
\caption{Root mean square error over 30 simulated networks ($\pm$ 2 standard errors) for different parameters}
\label{fig:sim1_supp}
\end{figure*}

The root mean squared error (RMSE) for the latent positions, each of the model parameters, and baseline intensities $\mu_{uv}$ for all node pairs is shown in Figure \ref{fig:sim1_supp}. 
As the time duration $T$ increases, more events are generated. 
Notice that the errors are all decreasing with increasing $T$ as one would expect.

\subsection{Real Network Dataset Descriptions}
\label{sec:supp_datasets}
We perform experiments on several real network datasets: 
\begin{itemize}
  \item Reality mining dataset \citep{eagle2006reality} is derived from the Reality Commons project from 14 September 2004 to 5 May 2005. The dataset we use contains 65 students and 2150 communications (Denote as Reality).
  \item Enron email dataset \citep{klimt2004enron} consists communications among 155 Enron individuals. It contains 9,646 email message spanning a period of 453 days. This dataset is tested in Yang's paper for their DLS model (Denote as Enron)
  \item The Militarized Interstate Disputes (MID) dataset \citep{palmer2021mid5} consists of 145 countries with 5088 Disputes among them with a period of 8380 days (Denote as MID). 
  \item Facebook-forum dataset \citep{nr} consists of 899 students posted 33,720 broadcast messages in the forum over 165 days (Denote as FB-forum).
\end{itemize}

For the Reality, FB-forum, and MID datasets, we re-scale the timestamps so that they are in all $[0, 1000]$ in the same manner as \citet{arastuie2020chip}. For Enron, we keep the same scale as \citet{yang2017decoupling} to make a fair comparison against the DLS model. 

\subsection{Descriptions of Other Models}
\label{sec:supp_models}

\paragraph{Dual Latent Space (DLS) Model}
We use the implementation at \url{https://github.com/jiaseny/lspp} for the DLS model. 
We provide a detailed comparison of the DLS model with our proposed LSH model in Section \ref{sec:relation_DLS}. 
It adopts a bivariate Hawkes process and latent space-based approach to capture homophily and reciprocity of continuous-time dynamic networks. Unlike our proposed LSH model, the DLS model does not have the self-excitation term. Their reciprocal terms are parameterized by multiple latent spaces associated with different decay values.

\paragraph{Hawkes Process-based Block Models}
We use the implementation at \url{https://github.com/IdeasLabUT/CHIP-Network-Model} for the Community Hawkes Independent Pairs (CHIP) model  \citep{arastuie2020chip} and the Block Hawkes Model (BHM) \citep{junuthula2019block}. 
CHIP is a univariate Hawkes process network model with block structure where each node pair is independent of all others. 
The BHM is also a univariate Hawkes process network model with block structure; however, an event between a node pair equally excites all node pairs in the same block pair.

\paragraph{Continuous-time Dynamic Network Embeddings (CTDNE)}
We used the same hyperparameters ($d=128, R=10, L=80, \omega=10$) as mentioned in \citet{nguyen2018continuous}. We used the implementation from the StellarGraph package \citep{StellarGraph}. We used timestamps up to $t_i$ (beginning of test window) to create both temporal walks and the classifier's positive and negative examples. Edges feature vector is computed using weighted-L2 operation. To create walks, we test using directed/undirected graphs, also varied the neighbor selection distribution between biased(exponential)/unbiased. Best results are reported for each dataset.

\subsection{Dynamic Link Prediction}
\label{sec:supp_dynamic_lp}

\begin{figure}[t]
\centering
\includegraphics[width=4in]{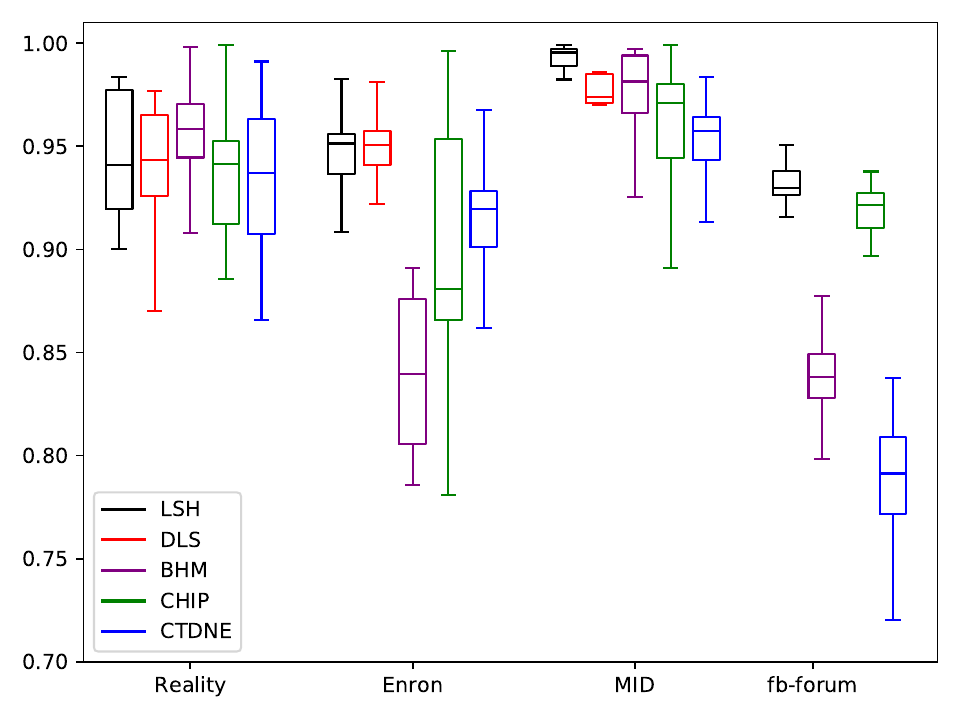}
\caption{AUC values for dynamic link prediction over 100 random time intervals}
\label{fig:dlp_boxplots}
\end{figure}

In section \ref{sec:real_experiment}, we perform dynamic link prediction experiments. In this section, we demonstrate more results of the AUC values and the ROC curves over 100 time intervals for different models and different real world datasets. 

Figure \ref{fig:dlp_boxplots} shows the boxplot of the AUC values for dynamic link prediction over 100 random time intervals. 
The DLS model does not scale to the fb-forum dataset. 
The box plot indicates that our proposed model achieves the best in MID and fb-forum dataset and is competitive in Reality and Enron datasets. Moreover, in the fb-forum, MID, and Enron datasets, the dispersion in AUC values from our model is less as evidenced by the low width of the boxes. In all datasets, the bulk of the distribution of AUC values for our LSH model is above 90\% indicating a superior performance in the dynamic link prediction task.

\begin{figure}[p]
\newcommand{\figwidth}{3.0in}
\centering 
    \begin{subfigure}[c]{\figwidth}
        \centering
        \hfill
        \includegraphics[width=\figwidth]{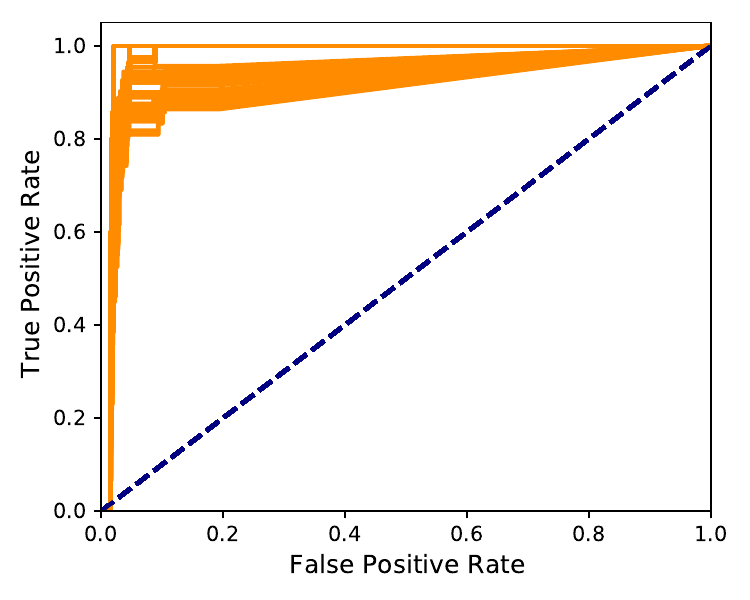}
        \caption{ROC curve of LSH model for Reality Mining}
    \end{subfigure}
    \begin{subfigure}[c]{\figwidth}
        \centering
        \hfill
        \includegraphics[width=\figwidth]{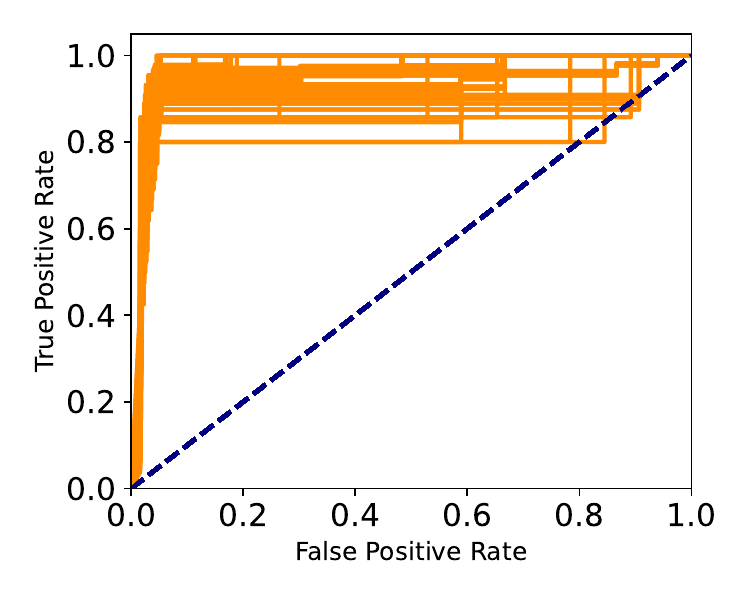}
        \caption{ROC curve of DLS model for Reality Mining}
    \end{subfigure}
    
    \begin{subfigure}[c]{\figwidth}
        \centering
        \hfill
        \includegraphics[width=\figwidth]{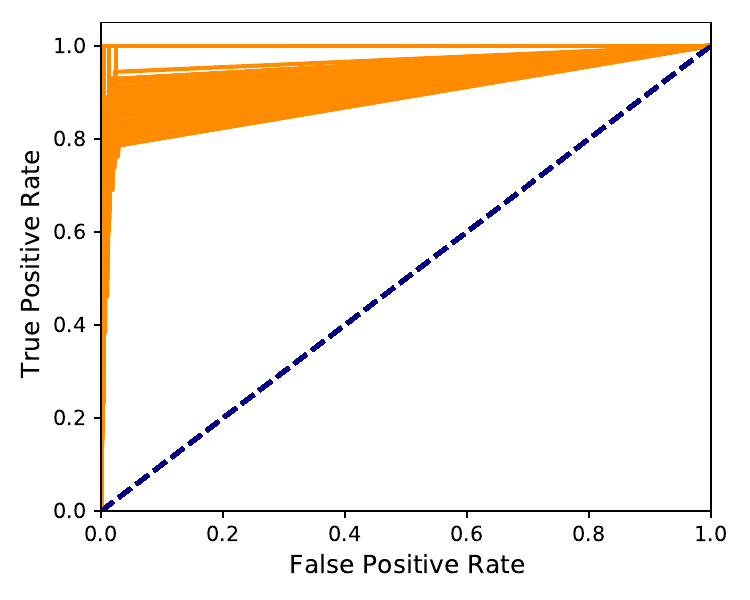}
        \caption{ROC curve of CHIP model for Reality Mining}
    \end{subfigure}
    \begin{subfigure}[c]{\figwidth}
        \centering
        \hfill
        \includegraphics[width=\figwidth]{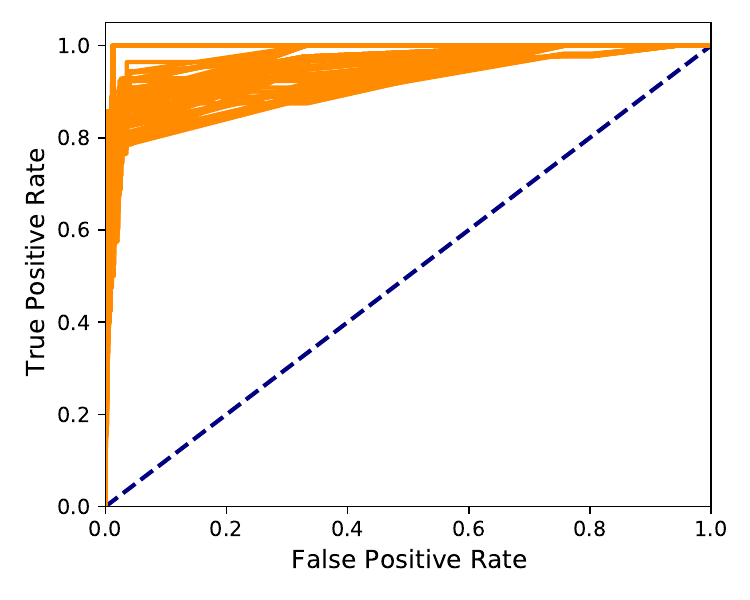}
        \caption{ROC curve of BHM model for Reality Mining}
    \end{subfigure}
    
\caption{Dynamic link prediction on 100 random time intervals on Reality Mining}
\label{fig:dlp_reality}
\end{figure}

\begin{figure}[p]
\centering 
    \begin{subfigure}[c]{3.0in}
        \newcommand{\figwidth}{3.0in}
        \centering
        \hfill
        \includegraphics[width=\figwidth]{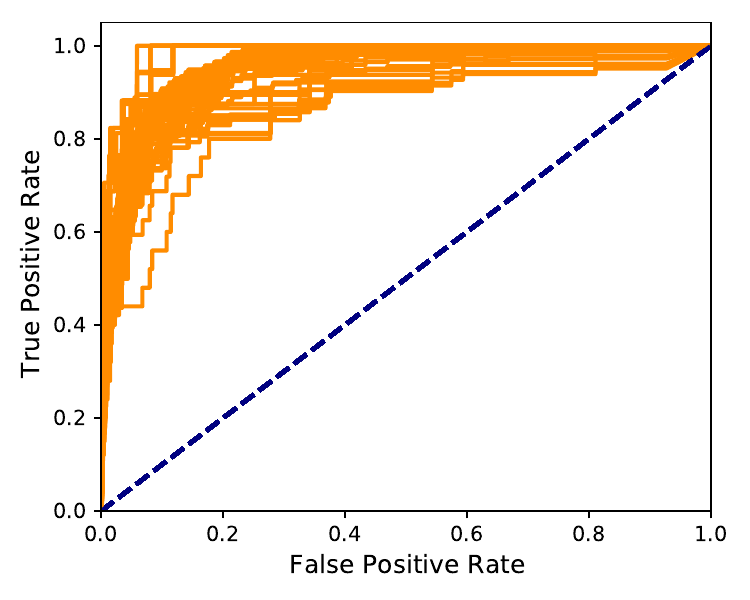}
        \caption{ROC curve of LSH model for Enron}
    \end{subfigure}
    \begin{subfigure}[c]{3.0in}
        \newcommand{\figwidth}{3.0in}
        \centering
        \hfill
        \includegraphics[width=\figwidth]{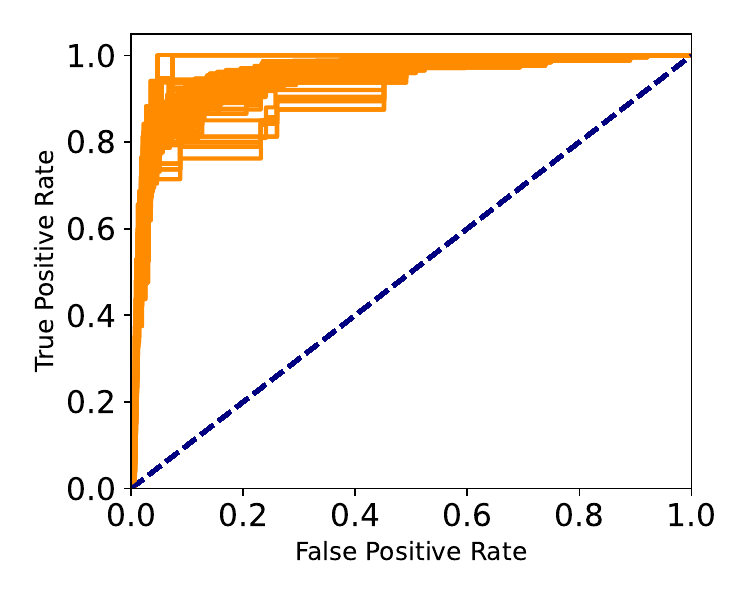}
        \caption{ROC curve of DLS model for Enron}
    \end{subfigure}
    
    \begin{subfigure}[c]{3.0in}
        \newcommand{\figwidth}{3.0in}
        \centering
        \hfill
        \includegraphics[width=\figwidth]{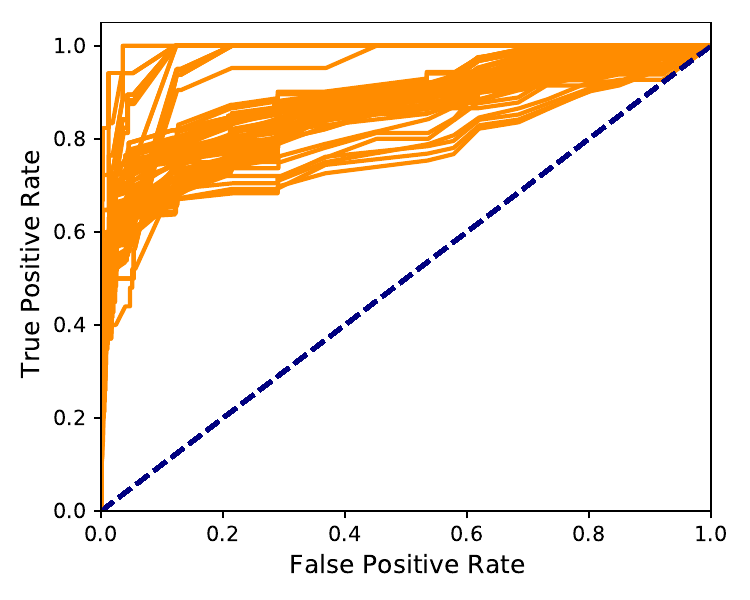}
        \caption{ROC curve of CHIP model for Enron}
    \end{subfigure}
    \begin{subfigure}[c]{3.0in}
        \newcommand{\figwidth}{3in}
        \centering
        \hfill
        \includegraphics[width=\figwidth]{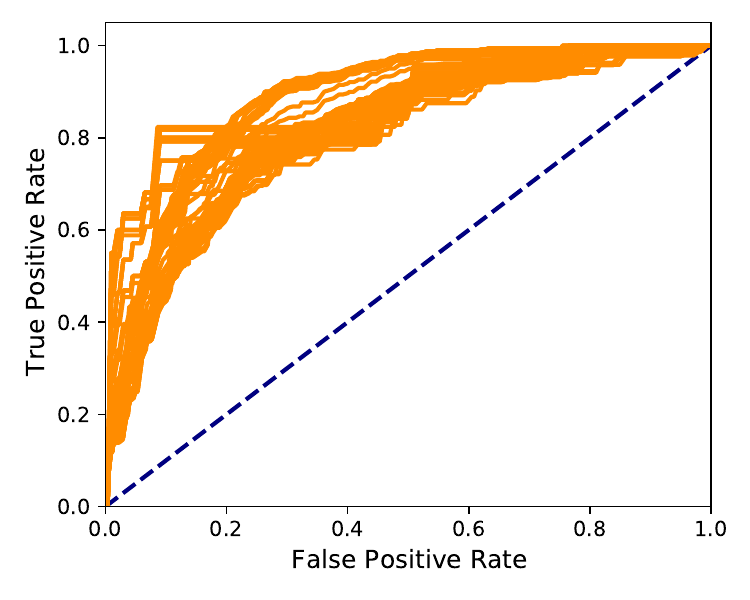}
        \caption{ROC curve of BHM model for Enron}
    \end{subfigure}
\caption{Dynamic link prediction on 100 random time intervals on Enron}
\label{fig:dlp_enron}
\end{figure}

\begin{figure}[p]
\centering 
    \begin{subfigure}[c]{3.0in}
        \newcommand{\figwidth}{3.0in}
        \centering
        \hfill
        \includegraphics[width=\figwidth]{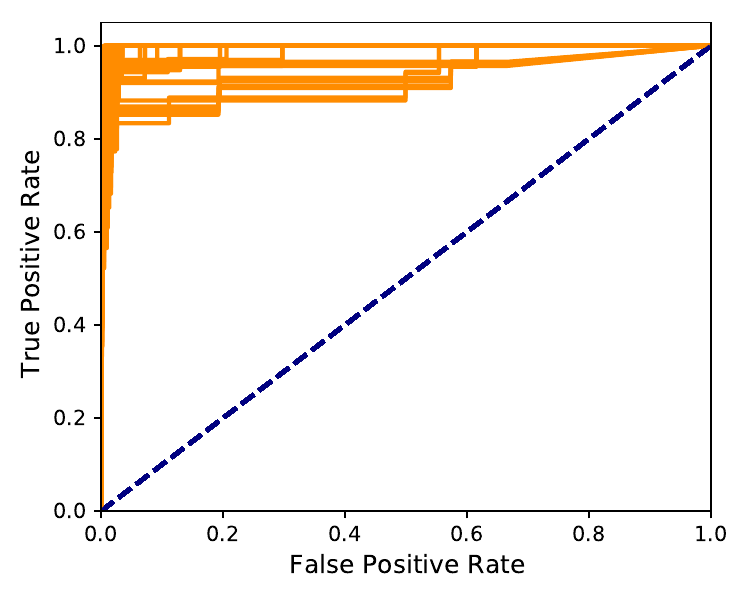}
        \caption{ROC curve of LSH model for MID}
    \end{subfigure}
    \begin{subfigure}[c]{3.0in}
        \newcommand{\figwidth}{3.0in}
        \centering
        \hfill
        \includegraphics[width=\figwidth]{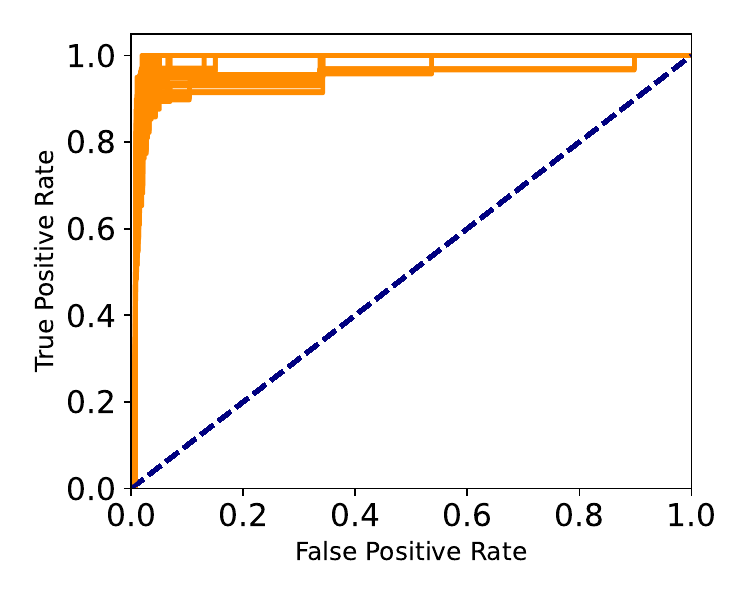}
        \caption{ROC curve of DLS model for MID}
    \end{subfigure}
    
    \begin{subfigure}[c]{3.0in}
        \newcommand{\figwidth}{3.0in}
        \centering
        \hfill
        \includegraphics[width=\figwidth]{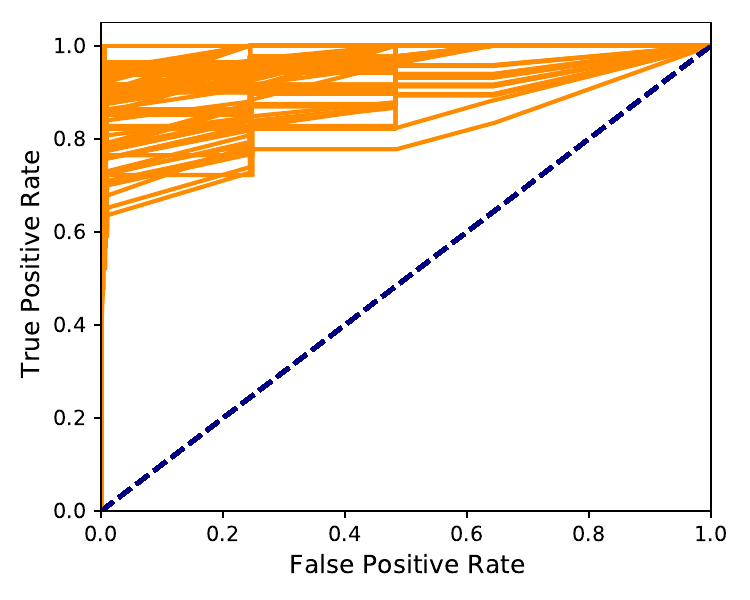}
        \caption{ROC curve of CHIP model for MID}
    \end{subfigure}
    \begin{subfigure}[c]{3.0in}
        \newcommand{\figwidth}{3.0in}
        \centering
        \hfill
        \includegraphics[width=\figwidth]{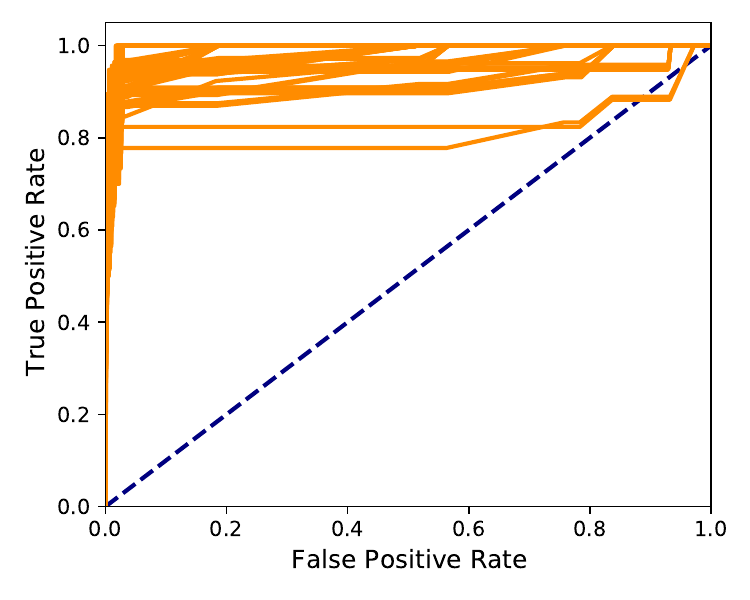}
        \caption{ROC curve of BHM model for MID}
    \end{subfigure}
\caption{Dynamic link prediction on 100 random time intervals on MID}
\label{fig:dlp_mid}
\end{figure}

\begin{figure}[p]
\centering 
    \begin{subfigure}[c]{3.0in}
        \newcommand{\figwidth}{3.0in}
        \centering
        \hfill
        \includegraphics[width=\figwidth]{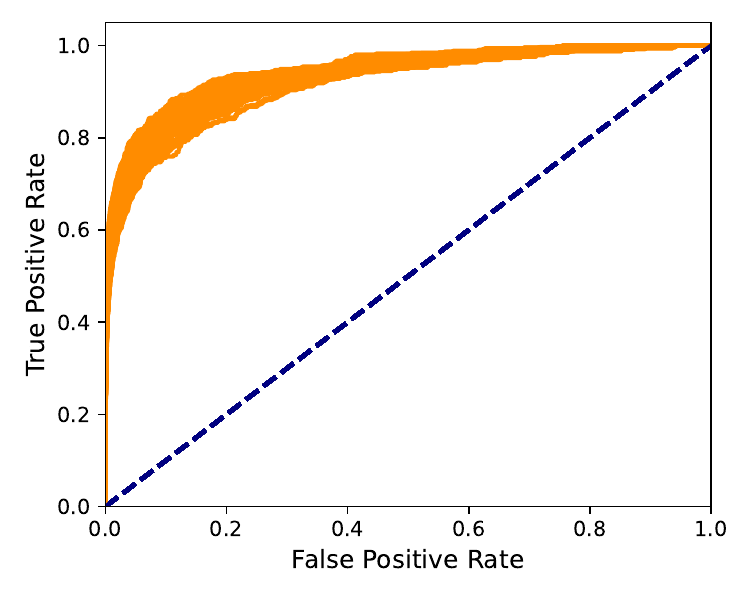}
        \caption{ROC curve of LSH model for fb-forum}
    \end{subfigure}

    \begin{subfigure}[c]{3.0in}
        \newcommand{\figwidth}{3.0in}
        \centering
        \hfill
        \includegraphics[width=\figwidth]{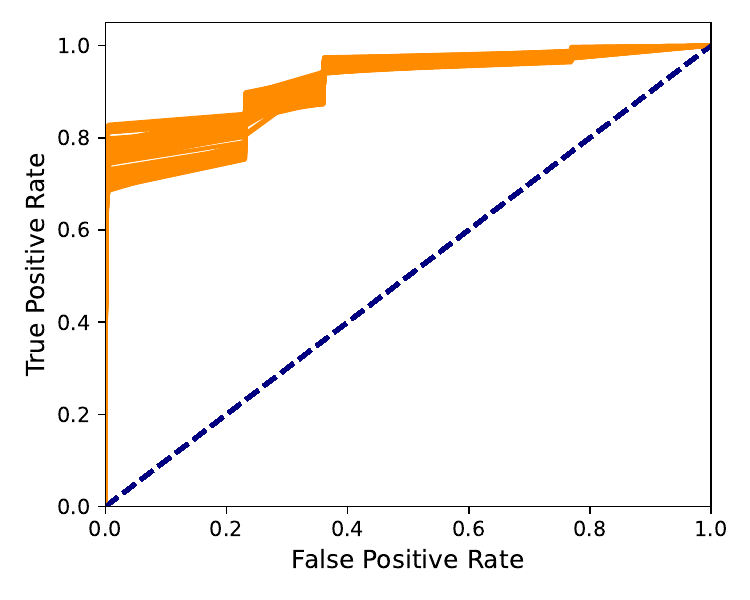}
        \caption{ROC curve of CHIP model for fb-forum}
    \end{subfigure}
    \begin{subfigure}[c]{3.0in}
        \newcommand{\figwidth}{3.0in}
        \centering
        \hfill
        \includegraphics[width=\figwidth]{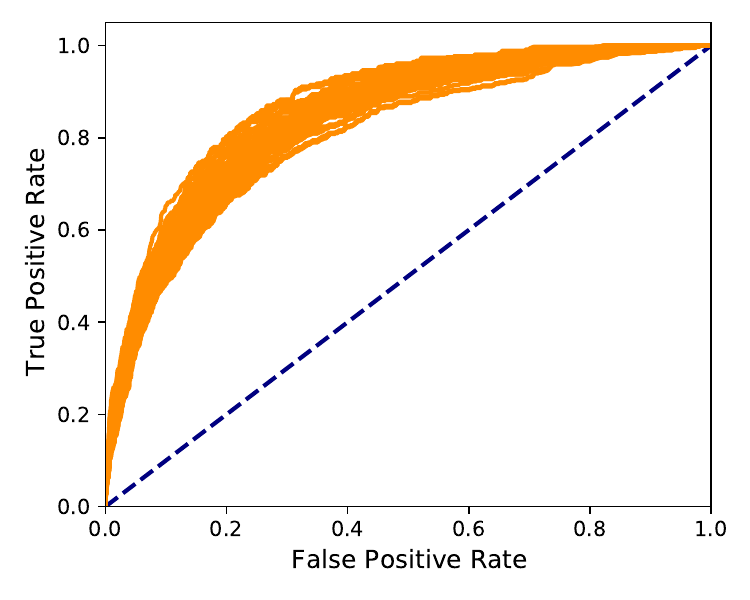}
        \caption{ROC curve of BHM model for fb-forum}
    \end{subfigure}
\caption{Dynamic link prediction on 100 random time intervals on fb-forum}
\label{fig:dlp_fb-forum}
\end{figure}

Figure \ref{fig:dlp_reality}-\ref{fig:dlp_fb-forum} demonstrate the corresponding ROC curves for different models and real world datasets. The curves matches what we observed in Table \ref{tab:predictive_accuracy} that our proposed model outperforms other models on MID and fb-forum and is competitive on Reality and Enron.

\subsection{Posterior Predictive Checks}
\label{sec:supp_ppc}
\begin{figure*}[p]
\centering 
    \begin{subfigure}[c]{3.0in}
        \newcommand{\figwidth}{3.0in}
        \centering
        \hfill
        \includegraphics[width=\figwidth]{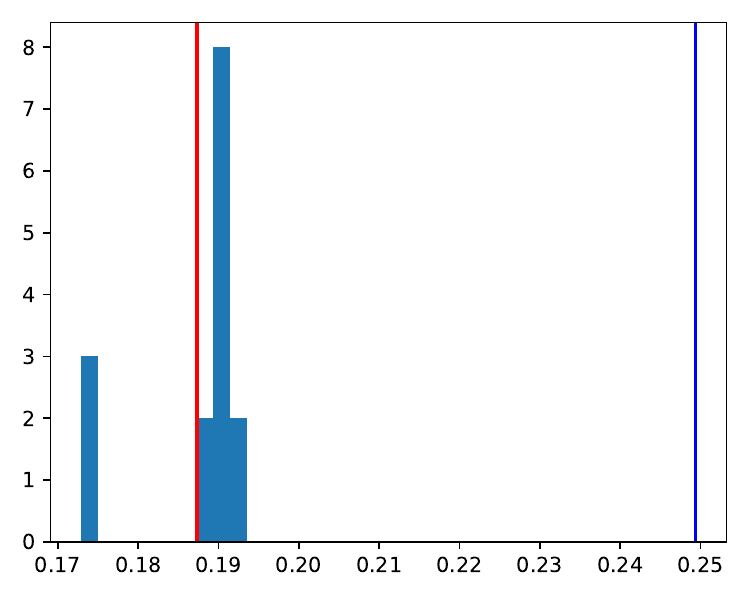}
        \caption{Histogram of average clustering coefficients}
    \end{subfigure}
    \begin{subfigure}[c]{3.0in}
        \newcommand{\figwidth}{3.0in}
        \centering
        \hfill
        \includegraphics[width=\figwidth]{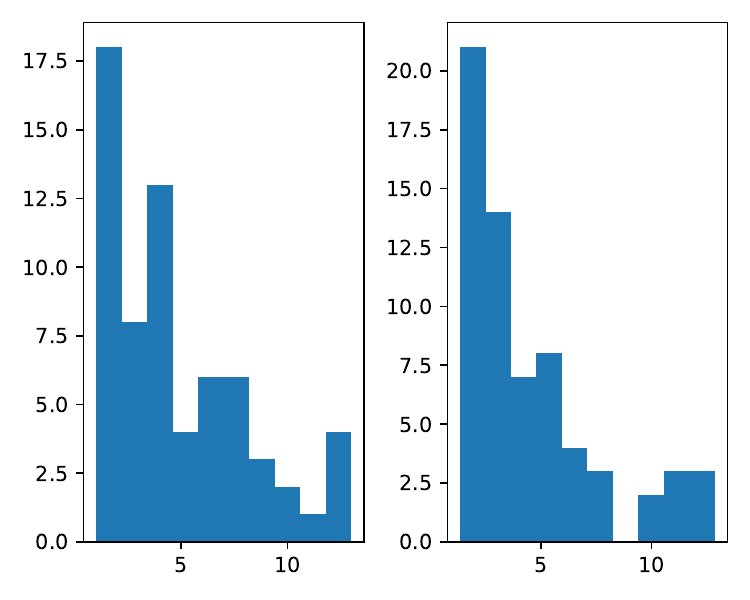}
        \caption{Histogram of degree distribution for real network(left) and mean of 15 simulated networks (right)}
    \end{subfigure}
    \begin{subfigure}[c]{3in}
        \newcommand{\figwidth}{3in}
        \centering
        \hfill
        \includegraphics[width=\figwidth]{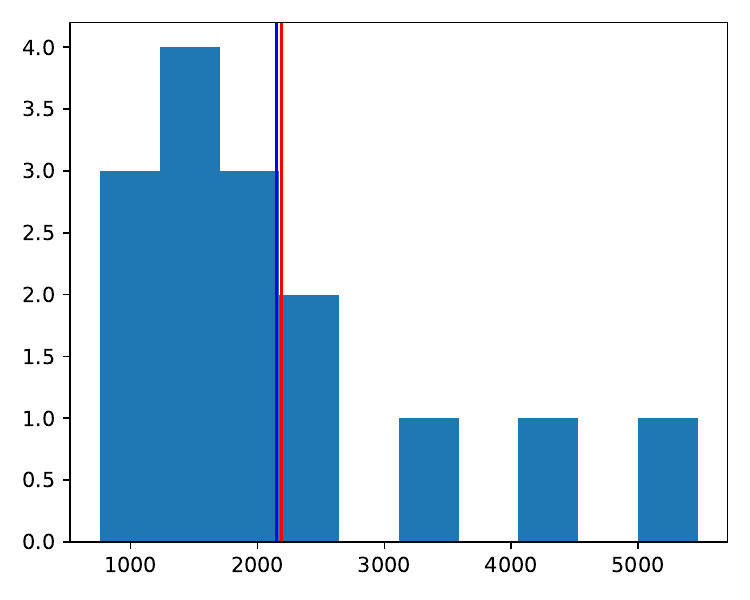}
        \caption{Histogram of number of counts}
    \end{subfigure}
        \begin{subfigure}[c]{3in}
        \newcommand{\figwidth}{3in}
        \centering
        \hfill
        \includegraphics[width=\figwidth]{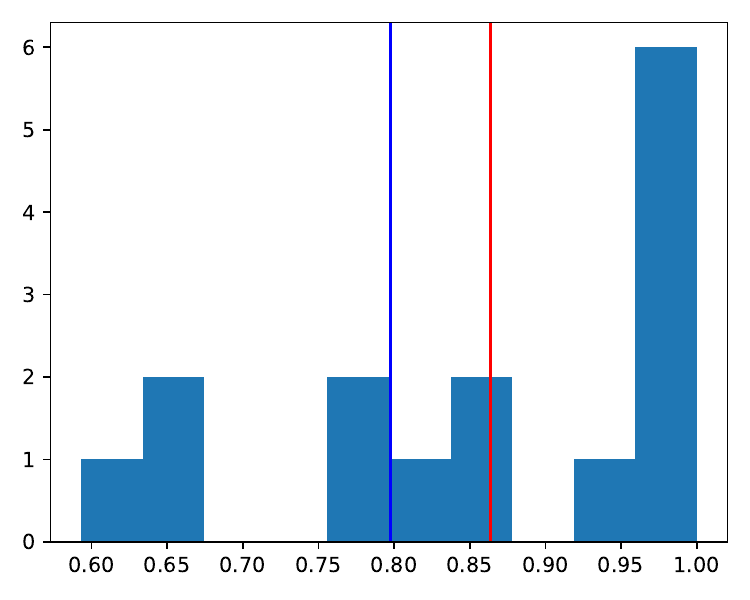}
        \caption{Histogram of reciprocity}
    \end{subfigure}
        \begin{subfigure}[c]{3in}
        \newcommand{\figwidth}{3in}
        \centering
        \hfill
        \includegraphics[width=\figwidth]{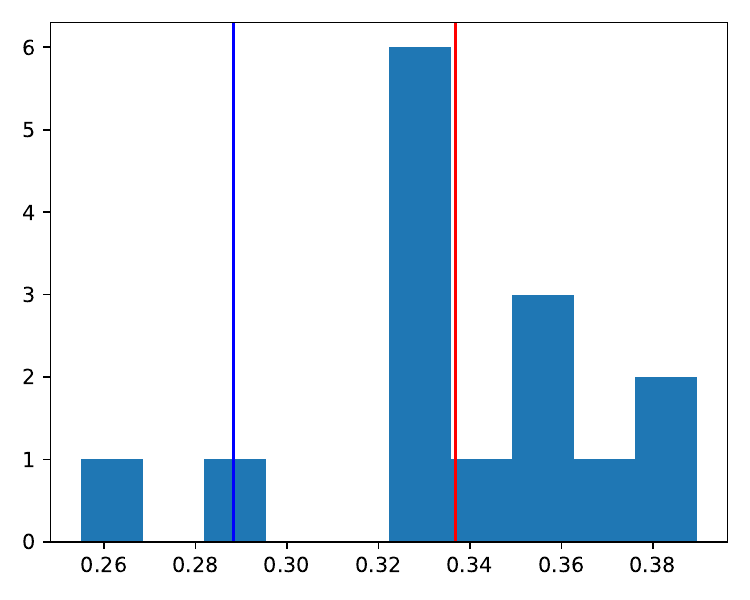}
        \caption{Histogram of transitivity}
    \end{subfigure}
\caption{Histogram of 15 simulations from the LSH model fitted to Reality mining (blue line: actual value; red line: mean of 15 simulated values).}
\label{fig:ppc_lsh}
\end{figure*}

\begin{figure*}[p]
\centering 
    \begin{subfigure}[c]{3.0in}
        \newcommand{\figwidth}{3.0in}
        \centering
        \hfill
        \includegraphics[width=\figwidth]{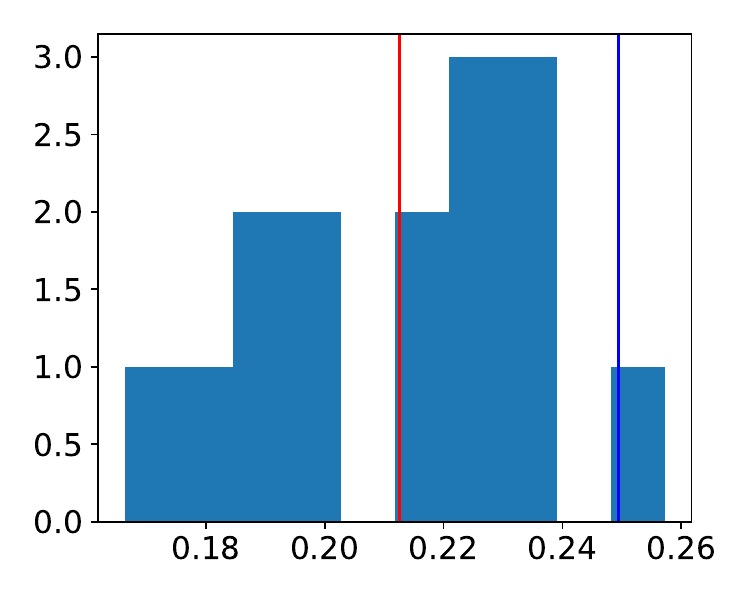}
        \caption{Histogram of average clustering coefficients}
    \end{subfigure}
    \begin{subfigure}[c]{3.0in}
        \newcommand{\figwidth}{3.0in}
        \centering
        \hfill
        \includegraphics[width=\figwidth]{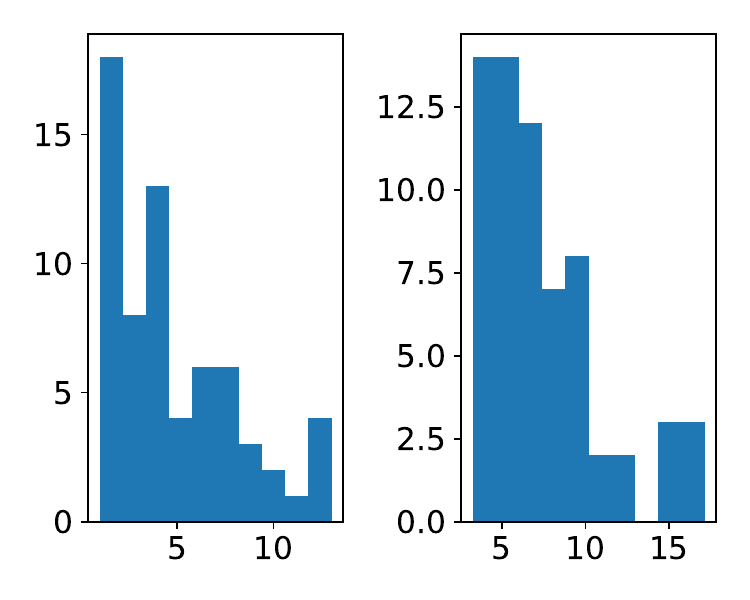}
        \caption{Histogram of degree distribution for real network(left) and mean of 15 simulated networks (right)}
    \end{subfigure}
    \begin{subfigure}[c]{3in}
        \newcommand{\figwidth}{3in}
        \centering
        \hfill
        \includegraphics[width=\figwidth]{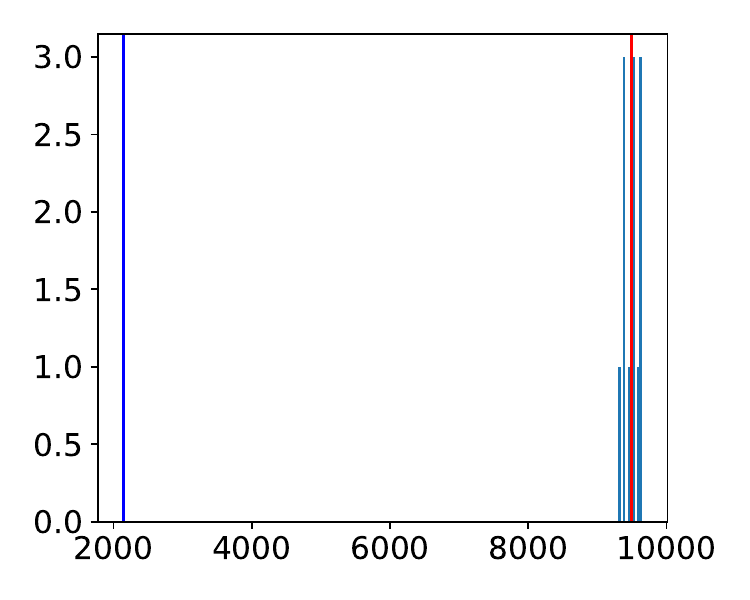}
        \caption{Histogram of number of counts}
    \end{subfigure}
        \begin{subfigure}[c]{3in}
        \newcommand{\figwidth}{3in}
        \centering
        \hfill
        \includegraphics[width=\figwidth]{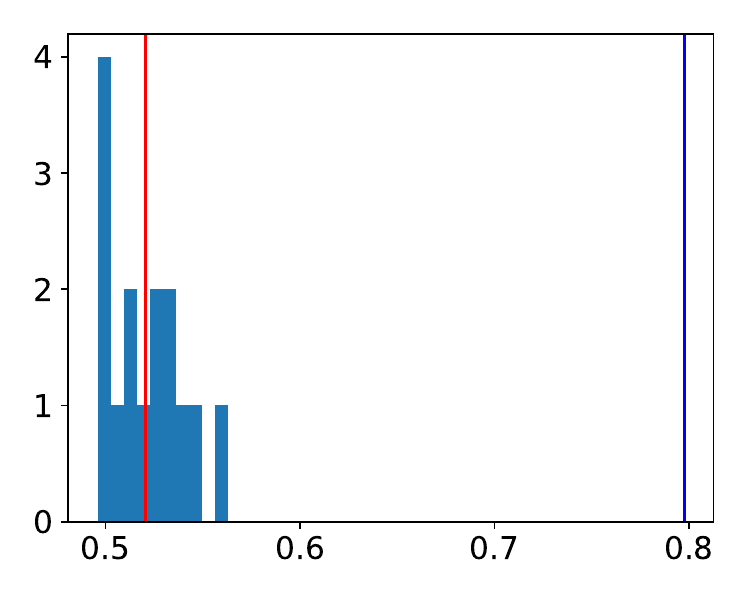}
        \caption{Histogram of reciprocity}
    \end{subfigure}
        \begin{subfigure}[c]{3in}
        \newcommand{\figwidth}{3in}
        \centering
        \hfill
        \includegraphics[width=\figwidth]{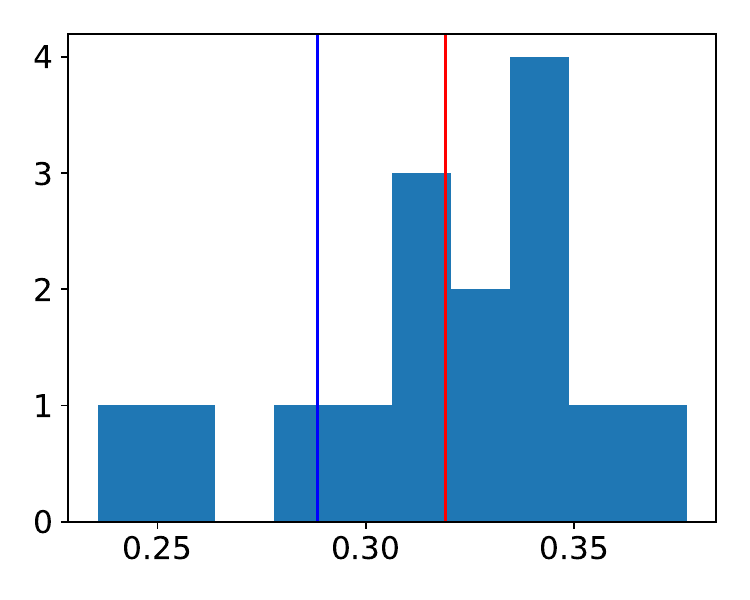}
        \caption{Histogram of transitivity}
    \end{subfigure}
\caption{Histogram of 15 simulations from the DLS model fitted to Reality mining (blue line: actual value; red line: mean of 15 simulated values).}
\label{fig:ppc_dls}
\end{figure*}

As we discussed in Section \ref{sec:generative_accuracy}, we simulate 15 networks from the fitted model and demonstrate the PPCs on the statistics in Table \ref{tab:generative_accuracy}. We show the corresponding histogram of the PPCs for Reality Mining in Figures \ref{fig:ppc_lsh} and \ref{fig:ppc_dls} for LSH and DLS, respectively. The actual value of the statistics is plotted as the blue vertical line. The red vertical line indicates the mean value of the statistics observed over the 15 simulated networks. Each figure consists of five subplots. The first, third, fourth, and fifth subplots give the histograms of average clustering coefficient, counts of events, reciprocity, and transitivity observed over the 15 simulated networks. The second subplot in each figure shows the histogram of the degree distribution. The left is the actual degree distribution, and the right is the degree distribution of the mean degree for each node of 15 simulated networks. 

In general, the LSH performs quite well on generating average clustering coefficients, number of events, reciprocity, transitivity, and degree distribution that are similar to what is observed in the corresponding real dataset. The DLS performs well on the average clustering coefficients, the transitivity, and the degree distribution but simulates orders of magnitude more events. The DLS model also fails to simulate the high reciprocity observed in the data.

\newpage
\section{Additional Case Study Results}
\label{sec:supp_case_study}

\begin{figure}[tp]
\centering 
\newcommand{\figwidth}{6.7in}
     \includegraphics[width=\figwidth]{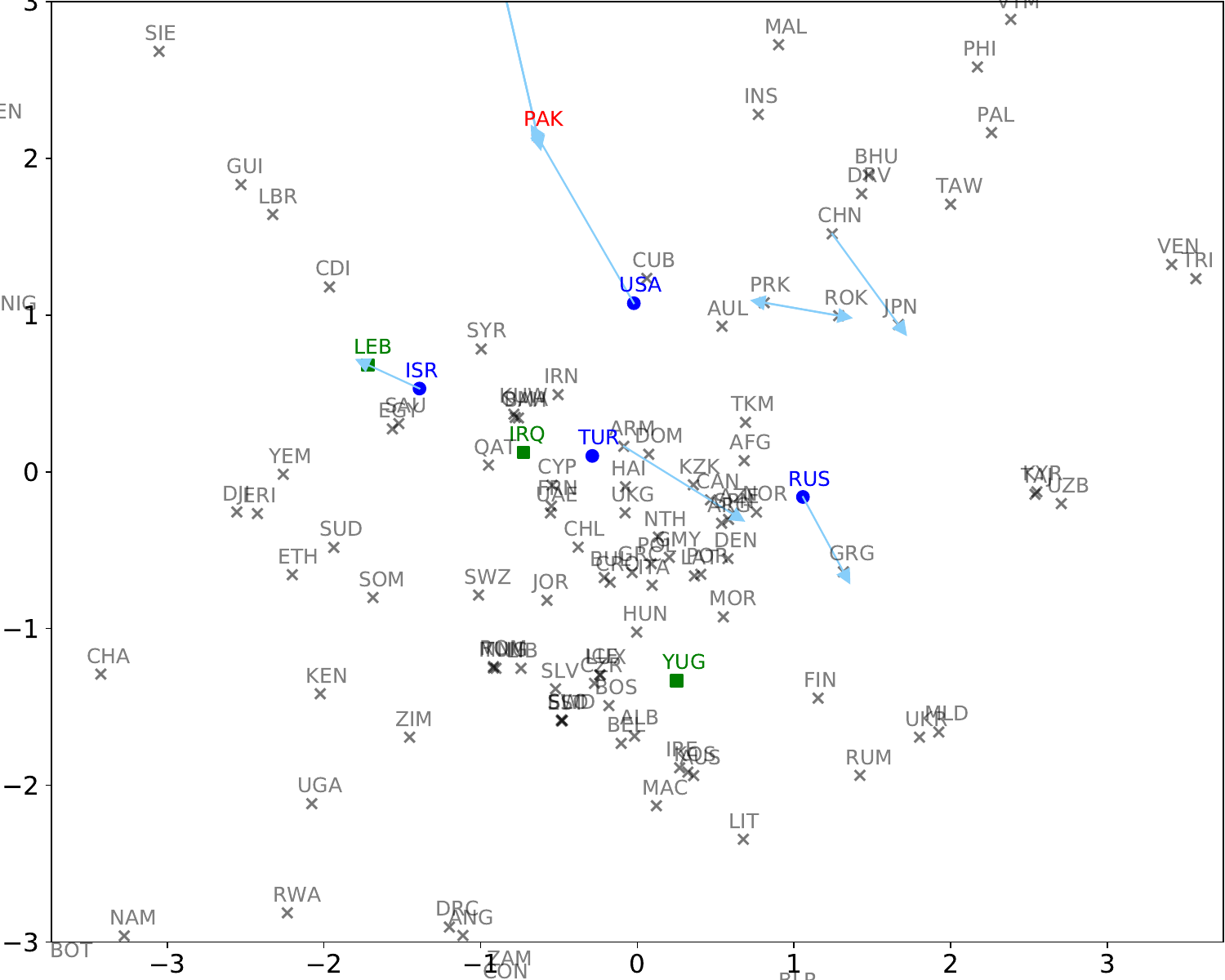}
\caption{2-D latent space plot for MID data with positive slope (zoomed in version of Figure \ref{fig:mid_latent_pos}).}
\label{fig:mid_latent_zoomed}
\end{figure}

\begin{figure}[tp]
\centering 
\newcommand{\figwidth}{6.7in}
     \includegraphics[width=\figwidth]{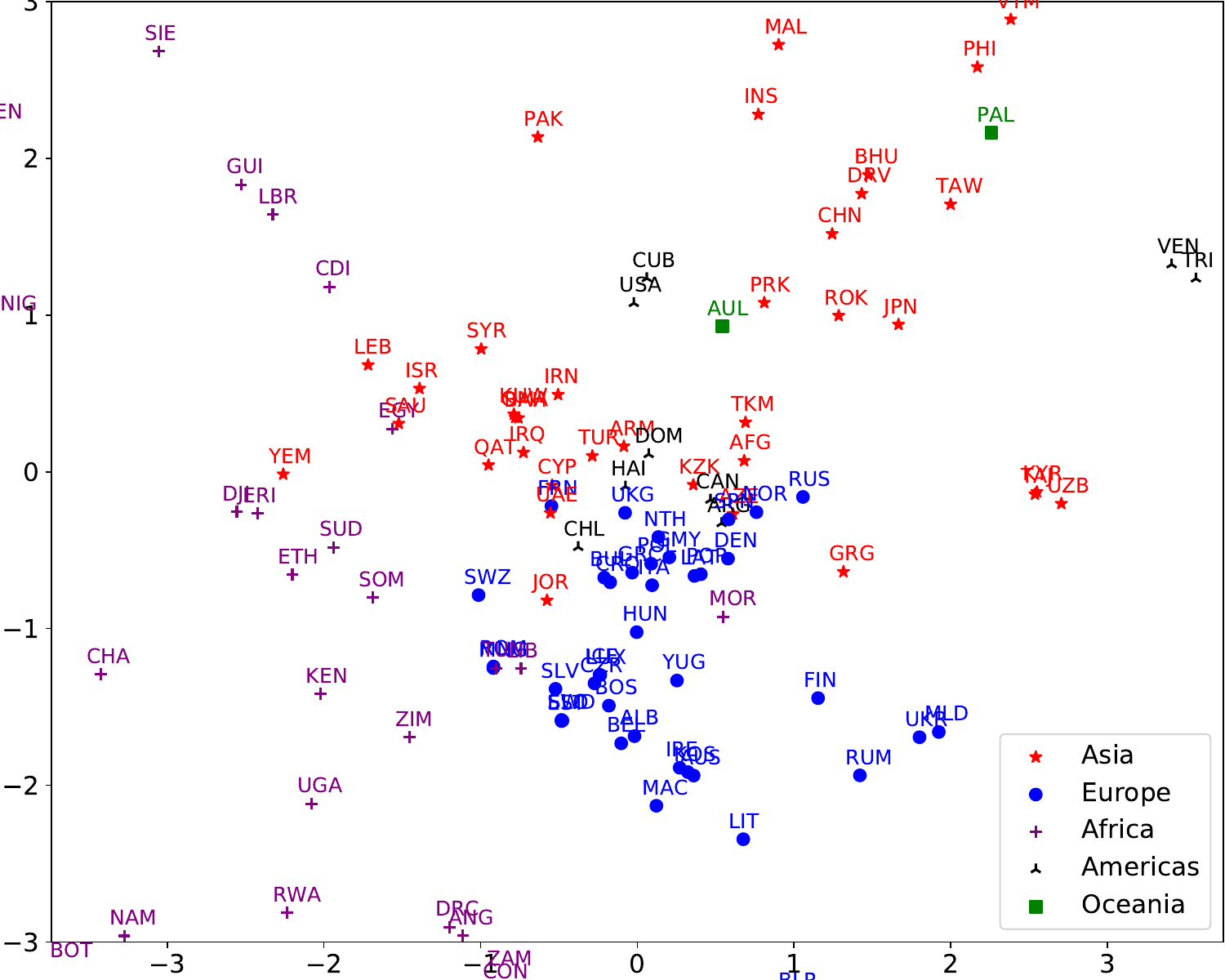}
\caption{2-D latent space plot for MID data with positive slope and countries colored by continent (zoomed in version of Figure \ref{fig:mid_latent_continents}).}
\label{fig:mid_latent_cont_zoomed}
\end{figure}

Figures \ref{fig:mid_latent_zoomed} and \ref{fig:mid_latent_cont_zoomed} show zoomed in versions of the latent space for the positive slope model shown in Figures \ref{fig:mid_latent_pos} and \ref{fig:mid_latent_continents}, respectively. 
In Table \ref{tab:mid_pair_incidence}, we show pairs of countries that have the top 50 number of incidents by one country towards another. 
In Figures \ref{fig:mid_latent} and \ref{fig:mid_latent_zoomed}, we draw edges to represent the top 10 frequently occurring incidence between pairs of countries in Table \ref{tab:mid_pair_incidence}. The edges with two-way arrows indicate that they serve both as initiator and receiver to the other (e.g.~PRK and ROK both have disputes with the other with the number of incidences 109 and 95, respectively).

Tables \ref{tab:mid_nodal_pos} and \ref{tab:mid_nodal_neg} show the estimated sender and receiver effects parameters from the LSH model with positive and negative slope, respectively. To compare with the highlighted countries in Figure \ref{fig:mid_latent} and \ref{fig:mid_latent_zoomed}, countries that initiate more incidents tend to have higher sender effect and countries that receive more incidents tend to have higher receiver effect, e.g.~USA has sender effect 4.18, and Yugoslavia (YUG) has receiver effect 3.03.

\begin{table}[tp]
    \caption{Pairs of countries that have the top 50 highest incidents}
    \centering
    \begin{tabular}{lll|lll}
    \hline
    Initiator  & Receiver & \# of incidents & Initiator & Receiver & \# of incidents \\ 
    \hline
    ISR  & LEB     & 588             & CHN  & TAW     & 38               \\
    PAK  & IND     & 174             & YUG  & ALB     & 37               \\
    USA  & IRQ     & 174             & SYR  & TUR     & 34               \\
    USA  & PAK     & 140             & ALB  & YUG     & 31               \\
    ARM  & AZE     & 128             & JPN  & CHN     & 31               \\
    IND  & PAK     & 116             & USA  & PRK     & 29               \\
    CHN  & JPN     & 110             & IRQ  & USA     & 29               \\
    PRK  & ROK     & 109             & UKG  & YUG     & 29               \\
    ROK  & PRK     & 95              & KUW  & IRQ     & 28               \\
    RUS  & GRG     & 88              & LEB  & ISR     & 28               \\
    TUR  & GRC     & 85              & FRN  & YUG     & 27               \\
    UKG  & IRQ     & 83              & AZE  & ARM     & 26               \\
    TUR  & IRQ     & 80              & RUS  & JPN     & 26               \\
    GRC  & TUR     & 73              & PHI  & CHN     & 26               \\
    SYR  & LEB     & 61              & NTH  & YUG     & 26               \\
    PAK  & AFG     & 61              & ITA  & YUG     & 26               \\
    TUR  & SYR     & 54              & CAM  & THI     & 25               \\
    TUR  & CYP     & 53              & TAW  & CHN     & 25               \\
    IRN  & IRQ     & 53              & SUD  & UGA     & 25               \\
    THI  & CAM     & 52              & GMY  & YUG     & 25               \\
    AFG  & PAK     & 51              & TUR  & YUG     & 25               \\
    ISR  & SYR     & 50              & BEL  & YUG     & 24               \\
    RUS  & AFG     & 44              & UKR  & RUS     & 24               \\
    IND  & BNG     & 44              & GRG  & RUS     & 24               \\
    USA  & YUG     & 39              & GRC  & YUG     & 24              
    \end{tabular}
    \label{tab:mid_pair_incidence}
\end{table}

\begin{table}[p]
    \caption{Nodal effect parameters estimated by the model with positive slope}
    \label{tab:mid_nodal_pos}
    \centering
    \begin{tabular}{lll|lll|lll|lll}
    \hline
       Country & send & receive & Country & send & receive & Country & send & receive & Country & send & receive \\ \hline
        AFG & -3.64 & 1.27 & DRV & -1.30 & -0.67 & LUX & -0.40 & -4.26 & SPN & 1.31 & -0.85 \\ 
        ALB & 0.34 & -0.11 & ECU & 2.41 & -1.91 & MAA & -3.89 & 2.14 & SRI & -0.08 & -1.67 \\ 
        ALG & 0.63 & -1.72 & EGY & 0.78 & 0.14 & MAC & -2.13 & 0.93 & SUD & 1.56 & 2.03 \\ 
        ANG & 0.81 & 0.69 & EQG & 2.07 & -2.21 & MAL & 0.89 & -4.16 & SUR & 1.75 & -0.94 \\ 
        ARG & 0.50 & -5.22 & ERI & 1.95 & 1.63 & MLD & -0.49 & 1.51 & SWD & -1.10 & -0.62 \\ 
        ARM & -0.24 & -1.25 & EST & -1.10 & -0.59 & MLI & 1.62 & 4.58 & SWZ & -4.77 & -0.10 \\ 
        AUL & 1.59 & -2.00 & ETH & 1.77 & 0.72 & MNG & -1.01 & -3.76 & SYR & 0.53 & 0.45 \\ 
        AUS & -4.54 & -0.48 & FIN & -4.57 & 0.24 & MOR & -0.97 & -1.71 & TAJ & 2.03 & 0.20 \\ 
        AZE & 0.10 & -0.44 & FRN & 1.44 & -0.51 & MYA & -3.80 & 5.48 & TAW & 1.62 & 1.35 \\ 
        BAH & -0.20 & -5.32 & GHA & 3.34 & 2.57 & NAM & 0.95 & 1.69 & TAZ & 0.49 & 2.61 \\ 
        BEL & 0.22 & 0.13 & GMY & 0.93 & -1.88 & NEP & -0.59 & -1.87 & THI & -0.71 & 5.22 \\ 
        BEN & -2.42 & -0.74 & GRC & 1.04 & -1.54 & NIC & 3.32 & 4.88 & TKM & -0.29 & -5.54 \\ 
        BHU & -4.13 & -1.45 & GRG & -4.07 & 1.20 & NIG & 5.07 & 0.91 & TOG & 2.51 & 1.24 \\ 
        BLR & 1.53 & -2.74 & GUI & 1.96 & -0.64 & NIR & 0.28 & 1.65 & TRI & -3.71 & 0.22 \\ 
        BNG & 0.24 & 2.57 & GUY & -2.08 & 1.76 & NOR & 1.17 & -1.04 & TUN & -1.01 & -3.70 \\ 
        BOS & -5.06 & -0.15 & HAI & -4.64 & 0.10 & NTH & 1.28 & -1.90 & TUR & 2.78 & -0.15 \\ 
        BOT & 0.36 & -2.67 & HON & 1.45 & 5.06 & OMA & -0.12 & -5.26 & UAE & 0.69 & -1.67 \\ 
        BRA & 1.48 & -0.80 & HUN & 0.35 & -4.19 & PAK & 2.60 & 2.28 & UGA & 1.50 & 3.01 \\ 
        BUI & 0.89 & 3.81 & ICE & -0.41 & -4.25 & PAL & -0.81 & -2.12 & UKG & 1.96 & -1.84 \\ 
        BUL & -0.26 & -0.92 & IND & 3.45 & 5.07 & PER & -1.27 & 2.49 & UKR & 0.88 & 2.62 \\ 
        CAM & -3.86 & 6.68 & INS & 1.78 & 0.67 & PHI & -0.57 & 2.52 & USA & 4.18 & 1.10 \\ 
        CAN & 1.81 & -1.28 & IRE & -4.26 & -0.67 & PNG & -1.65 & 1.71 & UZB & 2.75 & 0.95 \\ 
        CAO & 6.17 & -0.29 & IRN & 2.67 & 0.91 & POL & 0.33 & -1.41 & VEN & 2.45 & 1.92 \\ 
        CDI & 1.28 & 0.85 & IRQ & 0.82 & 2.53 & POR & 0.60 & -1.12 & VTM & 0.04 & 2.75 \\ 
        CEN & 3.80 & -4.27 & ISR & 2.19 & -0.13 & PRK & 2.00 & 0.68 & YEM & -3.84 & 0.63 \\ 
        CHA & 2.67 & 2.92 & ITA & 0.63 & -1.73 & QAT & 0.68 & -4.92 & YUG & 2.78 & 3.03 \\ 
        CHL & -4.97 & -1.65 & JAM & 1.47 & -1.14 & ROK & 1.30 & -0.10 & ZAM & -1.12 & 1.43 \\ 
        CHN & 3.76 & 1.53 & JOR & -0.31 & -1.25 & ROM & -0.98 & -3.77 & ZIM & -0.33 & 0.32 \\ 
        COL & 2.77 & 4.73 & JPN & 2.25 & 1.68 & RUM & -3.02 & 1.13 & ~ & ~ & ~ \\ 
        CON & -4.47 & 1.89 & KEN & 0.55 & 1.62 & RUS & 4.01 & 1.82 & ~ & ~ & ~ \\ 
        COS & -1.90 & 0.12 & KOS & -4.18 & -0.56 & RWA & 0.62 & 3.04 & ~ & ~ & ~ \\
        CRO & 0.61 & -0.45 & KUW & 0.17 & -5.34 & SAL & -4.19 & 2.05 & ~ & ~ & ~ \\ 
        CUB & 0.02 & -5.59 & KYR & 2.01 & 0.08 & SAU & 0.56 & -0.10 & ~ & ~ & ~ \\ 
        CYP & -4.62 & -1.33 & KZK & 0.20 & -5.64 & SIE & -1.30 & 2.40 & ~ & ~ & ~ \\ 
        CZR & -0.44 & -3.13 & LAT & -0.83 & -1.44 & SIN & -4.26 & 1.08 & ~ & ~ & ~ \\
        DEN & 1.09 & -1.59 & LBR & 1.94 & 1.34 & SLO & -1.11 & -0.60 & ~ & ~ & ~ \\ 
        DJI & -1.61 & 0.25 & LEB & -3.94 & -0.30 & SLV & -1.14 & -0.11 & ~ & ~ & ~ \\ 
        DOM & -0.35 & -5.66 & LIB & 0.30 & 2.93 & SOL & -4.57 & 3.22 & ~ & ~ & ~ \\ 
        DRC & 1.50 & 3.37 & LIT & 0.56 & 1.77 & SOM & -4.16 & 1.04 & ~ & ~ & ~ \\ 

    \end{tabular}
\end{table}

\begin{table}[p]
    \caption{Nodal effect parameters estimated by the model with negative slope}
    \label{tab:mid_nodal_neg}
    \centering
    \begin{tabular}{lll|lll|lll|lll}
    \hline
        Country & send & receive & Country & send & receive & Country & send & receive & Country & send & receive \\ \hline
        AFG & -1.49 & 3.09 & DRV & -1.50 & -0.16 & LUX & 0.43 & -1.59 & SPN & 1.58 & 1.20 \\ 
        ALB & 1.12 & -0.16 & ECU & 0.27 & -2.38 & MAA & -1.55 & -0.37 & SRI & -0.51 & -1.48 \\ 
        ALG & -0.49 & -2.00 & EGY & 0.43 & 0.44 & MAC & -1.34 & 0.65 & SUD & 1.37 & 1.89 \\ 
        ANG & 1.42 & -0.42 & EQG & -0.41 & -1.40 & MAL & 0.80 & -1.39 & SUR & -0.44 & -1.67 \\ 
        ARG & 1.07 & -1.38 & ERI & 1.11 & -0.54 & MLD & -0.79 & -0.25 & SWD & -0.52 & -0.38 \\ 
        ARM & -0.28 & 0.59 & EST & -0.45 & -0.53 & MLI & -0.41 & 0.46 & SWZ & -1.55 & 0.51 \\ 
        AUL & 1.39 & -0.28 & ETH & 0.88 & -0.25 & MNG & -0.66 & -1.42 & SYR & 0.34 & 0.95 \\ 
        AUS & -1.74 & -0.30 & FIN & -1.69 & 0.43 & MOR & -0.36 & -0.54 & TAJ & 0.55 & -0.41 \\ 
        AZE & 0.02 & 1.38 & FRN & 1.94 & 0.98 & MYA & -1.53 & 1.08 & TAW & 0.59 & 0.67 \\ 
        BAH & -0.37 & -1.60 & GHA & 0.43 & -1.40 & NAM & 0.46 & -0.27 & TAZ & -0.38 & -1.69 \\ 
        BEL & 0.76 & 0.61 & GMY & 1.38 & -0.26 & NEP & -0.42 & -1.39 & THI & 0.14 & 0.91 \\ 
        BEN & -1.63 & -0.32 & GRC & 1.59 & -0.21 & NIC & 0.46 & -0.22 & TKM & -0.36 & -1.41 \\ 
        BHU & -1.49 & -0.37 & GRG & -1.62 & 1.36 & NIG & 1.34 & -0.86 & TOG & -1.56 & -0.41 \\ 
        BLR & -0.52 & -1.40 & GUI & 0.82 & -1.57 & NIR & -0.55 & -0.42 & TRI & -2.01 & -0.30 \\ 
        BNG & 0.49 & -1.80 & GUY & -1.50 & 0.54 & NOR & 0.98 & 0.17 & TUN & -0.41 & -1.53 \\ 
        BOS & -1.48 & 0.48 & HAI & -1.55 & 1.85 & NTH & 1.64 & -0.24 & TUR & 2.60 & 1.34 \\ 
        BOT & -0.51 & -1.45 & HON & -0.51 & 0.37 & OMA & -0.21 & -1.41 & UAE & 1.15 & -0.45 \\ 
        BRA & -0.49 & -1.66 & HUN & 1.29 & -1.28 & PAK & 0.47 & 0.71 & UGA & 0.48 & 1.64 \\ 
        BUI & -0.37 & 0.54 & ICE & 0.50 & -1.37 & PAL & -0.37 & -1.62 & UKG & 2.18 & -0.29 \\ 
        BUL & 0.46 & 0.36 & IND & -0.26 & 1.43 & PER & -1.56 & 0.17 & UKR & -0.37 & 1.27 \\ 
        CAM & -1.75 & 0.54 & INS & 1.26 & 1.25 & PHI & -0.36 & 1.51 & USA & 3.04 & 1.74 \\ 
        CAN & 2.13 & 0.60 & IRE & -1.66 & -0.29 & PNG & -0.36 & -0.32 & UZB & 0.95 & -1.35 \\ 
        CAO & -0.45 & -0.75 & IRN & 1.78 & 2.12 & POL & 0.92 & -0.23 & VEN & 0.93 & -0.19 \\ 
        CDI & 0.50 & 1.25 & IRQ & -0.17 & 3.36 & POR & 1.06 & 0.62 & VTM & 0.30 & 0.99 \\ 
        CEN & 0.30 & -1.38 & ISR & 2.00 & -0.26 & PRK & 0.59 & 1.69 & YEM & -1.48 & 0.57 \\ 
        CHA & 1.13 & 1.30 & ITA & 1.19 & -0.41 & QAT & 1.23 & -1.36 & YUG & 3.05 & 3.66 \\ 
        CHL & -1.49 & -0.26 & JAM & -0.53 & -1.68 & ROK & 0.47 & 0.75 & ZAM & -1.48 & 0.45 \\ 
        CHN & 2.81 & 1.04 & JOR & 0.35 & -0.26 & ROM & -0.36 & -1.85 & ZIM & 0.56 & -0.45 \\ 
        COL & -1.44 & 0.86 & JPN & -0.38 & 1.60 & RUM & -1.29 & 0.50 & ~ & ~ & ~ \\ 
        CON & -1.94 & 0.61 & KEN & 0.35 & -0.26 & RUS & 3.40 & 2.59 & ~ & ~ & ~ \\ 
        COS & -0.39 & -1.42 & KOS & -1.49 & -0.45 & RWA & -0.38 & 1.63 & ~ & ~ & ~ \\ 
        CRO & 1.27 & 0.96 & KUW & -0.29 & -1.38 & SAL & -1.72 & -0.53 & ~ & ~ & ~ \\ 
        CUB & -0.50 & -1.44 & KYR & -1.27 & 0.56 & SAU & 0.49 & 0.56 & ~ & ~ & ~ \\ 
        CYP & -1.55 & -0.22 & KZK & 0.36 & -1.71 & SIE & -1.64 & 1.18 & ~ & ~ & ~ \\ 
        CZR & 0.32 & -1.29 & LAT & -0.54 & -0.22 & SIN & -1.55 & -0.29 & ~ & ~ & ~ \\ 
        DEN & 1.39 & -0.11 & LBR & 0.30 & 1.02 & SLO & -0.51 & -0.38 & ~ & ~ & ~ \\ 
        DJI & -1.62 & -0.38 & LEB & -1.50 & -0.59 & SLV & -0.43 & 0.49 & ~ & ~ & ~ \\ 
        DOM & -0.28 & -1.72 & LIB & 0.55 & 3.34 & SOL & -1.51 & -0.34 & ~ & ~ & ~ \\ 
        DRC & 1.58 & 0.75 & LIT & 0.49 & 1.08 & SOM & -1.49 & 1.05 & ~ & ~ & ~ \\ 
    \end{tabular}
\end{table}

\end{document}